\documentclass[lettersize,journal]{IEEEtran}

\usepackage{amsthm}
\usepackage{amsmath}    % 数学公式
\usepackage{amssymb}    % 数学符号
\usepackage{bm}         % 加粗数学符号
\usepackage{cite}       % 引用格式
\usepackage{graphicx}   % 图片支持
\usepackage{algorithm}
\usepackage{algpseudocode}
\usepackage{amsfonts}
\usepackage{multirow} % 用于多行合并单元格
\usepackage{subcaption}
\usepackage{bbm}

\begin{document}
	
	% 文章标题
	\title{Multi-Objective Communication Optimization for Temporal Continuity in Dynamic Vehicular Networks}
	
	% 作者和作者单位
	\author{Weian Guo, Wuzhao Li, Li Li, Lun Zhang and Dongyang Li
		% <-this % stops a space
		% <-this % stops a space
		\thanks{This work was supported in part by the National Natural Science Foundation of China under Grant 62273263, 72171172, 92367101 and 71771176; the Aeronautical Science Foundation of China under Grant 2023Z066038001; the National Natural Science Foundation of China Basic Science Research Center Program under Grant 62088101;Municipal Science and Technology Major Project (2022-5-YB-09); Natural Science Foundation of Shanghai under Grant Number 23ZR1465400. (Corresponding author: Wuzhao Li). Thanks for Mr. Wanli CAI for providing experiment environments to debug the codes.}
		\thanks{Weian Guo and Dongyang Li are with Sino-German College of Applied Sciences, Tongji University, Shanghai, China (email:\{guoweian, lidongyang0412\}@163.com). Wuzhao LI is with the Yongjia College, Wenzhou Polytechnic, Wenzhou, Zhejiang, China. Email:\{lwzhao055@wzpt.edu.cn\}. Lun Zhang is with School of Transportation, Tongji University, Shanghai, China (email: lun$\_$zhang@tongji.edu.cn).  Li Li is with the Department of Electronics and Information Sciences, Tongji University, Shanghai, China. Email:\{lili@tongji.edu.cn\}.}
	}
	
	% 制作标题
	\maketitle
	
	% 摘要
	\begin{abstract}
		Vehicular Ad-hoc Networks (VANETs) operate in highly dynamic environments characterized by high mobility, time-varying channel conditions, and frequent network disruptions. Addressing these challenges, this paper presents a novel temporal-aware multi-objective robust optimization framework, which for the first time formally incorporates temporal continuity into the optimization of dynamic multi-hop VANETs. The proposed framework simultaneously optimizes communication delay, throughput, and reliability, ensuring stable and consistent communication paths under rapidly changing conditions. A robust optimization model is formulated to mitigate performance degradation caused by uncertainties in vehicular density and channel fluctuations. To solve the optimization problem, an enhanced Non-dominated Sorting Genetic Algorithm II (NSGA-II) is developed, integrating dynamic encoding, elite inheritance, and adaptive constraint handling to efficiently balance trade-offs among conflicting objectives. Simulation results demonstrate that the proposed framework achieves significant improvements in reliability, delay reduction, and throughput enhancement, while temporal continuity effectively stabilizes communication paths over time. This work provides a pioneering and comprehensive solution for optimizing VANET communication, offering critical insights for robust and efficient strategies in intelligent transportation systems.
	\end{abstract}

	% 关键词
	
	\section{Introduction}
	\label{sec:intro}
	\IEEEPARstart{V}ehicular Ad-hoc Networks (VANETs) are a fundamental component of Intelligent Transportation Systems (ITS), enabling efficient communication among vehicles, roadside units (RSUs), and other infrastructure elements \cite{maram_bani_younes_59aa1648}. These networks support a wide range of applications, such as traffic management, accident prevention, and autonomous driving, by facilitating real-time information exchange \cite{eshita_rastogi_b566e6c6}. However, the highly dynamic nature of VANETs, characterized by rapid vehicular mobility, constantly changing network topologies, and time-varying channel conditions, poses significant challenges in maintaining reliable and efficient communication \cite{craig_cooper_aa6ee789}.
	
	The communication performance in VANETs is heavily influenced by factors such as network density, vehicular mobility, and the quality of the communication channel \cite{craig_cooper_aa6ee789}. The high mobility of vehicles often results in frequent topology changes, leading to increased packet loss and communication delays. In addition, uncertainties arising from varying vehicular density, weather conditions, and road types further complicate the task of maintaining stable communication paths. Therefore, achieving optimal communication performance in VANETs requires addressing both the dynamic nature of the network and the uncertainties induced by fluctuating traffic and channel conditions.
	
	Recent studies have proposed various optimization techniques to tackle these challenges, including routing protocols, channel allocation strategies, and network performance models \cite{junjie_zhang_fe5a8711}. However, many of these approaches focus on optimizing a single objective, such as communication delay or throughput, and do not account for the evolving nature of vehicular mobility over time \cite{md__noor_a_rahim_4b96d3b0, xianghui_cao_b8d3413d, wantanee_viriyasitavat_f5df62bb, craig_cooper_aa6ee789}. For example, traditional routing protocols often need to completely reconfigure communication paths when network topology changes, leading to significant overhead and temporary service disruptions. These methods typically fail to adapt to future topology changes, and performing real-time optimizations for each moment introduces significant computational overhead, potentially delaying the acquisition of optimal solutions. Furthermore, few existing methods explicitly address the robustness of communication paths under uncertain scenarios, which is crucial for ensuring reliable communication amidst mobility-induced network changes \cite{kai_xiong_ee6b167b}.
	
	Unlike existing approaches that focus solely on instantaneous performance, our framework uniquely addresses both current optimization needs and temporal stability. We present a multi-objective robust optimization framework for dynamic communication in VANETs, which incorporates temporal continuity to handle the evolving nature of vehicular mobility. The proposed method simultaneously optimizes three critical objectives—communication delay, throughput, and reliability—while ensuring stable and continuous communication paths over time as the network topology evolves. To address uncertainties such as fluctuating vehicle density and time-varying channel conditions, a robust optimization model is introduced to minimize performance degradation under uncertain scenarios. The optimization problem is solved using an enhanced version of the Non-dominated Sorting Genetic Algorithm II (NSGA-II), which is designed to incorporate temporal continuity and effectively balance trade-offs among competing objectives. The key contributions of this paper are as follows:
	\begin{itemize}
		\item A comprehensive multi-objective optimization framework that simultaneously optimizes communication delay, throughput, and reliability, explicitly addressing the dynamic nature of VANETs. Unlike existing methods, this framework incorporates long-term stability to ensure robust and consistent communication paths over time.
		\item The incorporation of temporal continuity, formally introduced in the optimization of dynamic VANETs for the first time. This innovation improves the adaptability of the network to frequent topology changes, significantly reducing the risk of communication disruptions and maintaining persistent communication paths in evolving environments.
		\item The development of a optimization model that minimizes performance degradation under uncertain conditions, such as fluctuating vehicle density and time-varying channel quality. Coupled with an enhanced Non-dominated Sorting Genetic Algorithm II (NSGA-II) featuring dynamic encoding and adaptive constraint handling, this model ensures reliable communication in highly variable scenarios.
	\end{itemize}
	
	The remainder of this paper is organized as follows: Section \ref{sec:related_work} reviews the related work on VANET communication optimization and robust optimization techniques. Section \ref{sec:system_model} presents the proposed multi-objective robust optimization framework. Section \ref{sec:algorithm} discusses the solution methodology and the enhanced NSGA-II algorithm used to solve the optimization problem. Section \ref{sec:sim} presents the simulation setup and results. Finally, Section \ref{sec:con} concludes the paper and discusses future research directions.

	\section{Related Work}
	\label{sec:related_work}
	The optimization of communication in Vehicular Ad-Hoc Networks (VANETs) has been extensively studied, with a wide range of methods proposed to address the dynamic and uncertain nature of these networks \cite{huacheng_zeng_7b63c042, francesco_malandrino_46ebdf5f, jiawen_kang_945bf2ac}. Most existing studies focus on optimizing key performance metrics, such as communication delay, throughput, and reliability \cite{zhonghui_pei_90e7494f, kai_xiong_ee6b167b, jin_tian_83309c8b}. However, a critical challenge remains unresolved: ensuring the long-term stability and temporal continuity of communication paths, which is essential in highly dynamic VANET environments \cite{jeng_ji_huang_855b9d66, felipe_d__cunha_945016f9}.
	
	\subsection{Optimization Techniques in VANETs}
	VANET communication optimization methods primarily address the challenges posed by frequent topology changes and high vehicular mobility \cite{silviu_andrei_lazar_493c4ad6, ralf_schmitz_f52c4b83, amina_bengag_31f1a673}. Routing protocols dominate this research area, with many studies proposing algorithms to identify optimal communication paths in real time by analyzing current network topology \cite{antonio_russoniello_44f40609, jiayue_he_ed1c7260}. While these methods are effective in reducing immediate communication delay and enhancing throughput, they often neglect the evolving nature of the network, leading to suboptimal performance over time \cite{xiaoyun_xie_85881fa0, lei_zhang_bfc944ca}. Additionally, dynamic channel allocation techniques, which adjust transmission power and frequency based on real-time conditions, have shown promise in improving throughput and reducing interference \cite{yasar_sinan_nasir_9cb0d554, linlin_sun_b468e9a7}. However, like routing protocols, these methods fail to account for the broader impact of changing vehicular mobility and long-term network evolution.
	
	While these optimization techniques address immediate network conditions, they fall short in achieving persistent communication paths that can adapt to the dynamic characteristics of VANETs. This limitation underscores the need for frameworks that incorporate temporal continuity and long-term stability into their design.
	
	\subsection{Robust Optimization and Multi-Objective Approaches}
	To mitigate uncertainties such as fluctuating vehicular density and varying channel conditions, robust optimization methods have been widely explored \cite{anirudh_subramanyam_78638638, kai_xiong_ee6b167b}. These methods aim to ensure reliable communication under uncertain scenarios by optimizing specific parameters, such as routing paths or link reliability \cite{jingqiu_guo_7aa54d64, chuan_xu_39cb2332}. However, most robust optimization approaches focus on individual performance metrics and overlook the impact of evolving network topologies over time \cite{alejandro_cohen_cafa02ff, nikoletta_sofra_c3fa0469}.
	
	Multi-objective optimization (MOO) has emerged as a powerful approach to balancing competing performance metrics such as delay, throughput, and reliability \cite{hamdy_h__el_sayed_48e78f4e}. While some studies extend MOO methods to robust optimization, they often fail to incorporate temporal continuity as a formal criterion. For example, many MOO-based routing protocols optimize delay and throughput without addressing the trade-offs introduced by additional objectives like reliability \cite{ke_li_f6a98174, tarik_taleb_4e090d42}. Consequently, these methods are limited in their ability to maintain stable communication paths over time.
	
	By integrating robust optimization with multi-objective frameworks, this paper advances beyond existing methods. Unlike previous works, our framework explicitly incorporates temporal continuity to ensure that communication paths remain stable and reliable despite dynamic changes in vehicular density and channel conditions.
	
	\subsection{Temporal Continuity in Dynamic VANETs}
	Temporal continuity, defined as the ability to maintain stable communication paths over time, is a crucial yet underexplored aspect of VANET optimization \cite{anirudh_paranjothi_116e5301, min_li_7968a175}. Frequent topology changes in VANETs disrupt communication paths, causing increased packet loss and delays \cite{yutong_liu_4f29ab21, azzedine_boukerche_3b82d91d}. While some studies propose routing algorithms that incorporate short-term stability by predicting future network conditions, these methods typically focus on immediate performance improvements and fail to address long-term stability \cite{mao_ye_30a7a94b, francesco_malandrino_46ebdf5f}.
	
	In contrast, the framework proposed in this paper formalizes temporal continuity as a key optimization objective. By doing so, it ensures that communication paths adapt not only to current conditions but also to the evolving nature of the network, providing long-term stability and improved performance in highly dynamic environments.
	
	\subsection{Summary of Related Work}
	Existing methods in VANET optimization have made significant progress in areas such as routing, channel allocation, and robust optimization \cite{zhonghui_pei_90e7494f}. However, most approaches focus on short-term improvements or single-objective optimization, neglecting the long-term stability of communication paths and the importance of temporal continuity. Robust optimization methods address uncertainties but often fail to consider the dynamic evolution of vehicular mobility, while multi-objective approaches rarely integrate temporal continuity into their frameworks.
	
	This paper bridges these gaps by proposing a novel multi-objective robust optimization framework that integrates temporal continuity. The proposed approach simultaneously optimizes delay, throughput, and reliability while leveraging an enhanced Non-dominated Sorting Genetic Algorithm II (NSGA-II) to adapt to dynamic and uncertain VANET environments. By incorporating temporal continuity as a formal criterion, this framework ensures stable and reliable communication paths over time, addressing the limitations of existing methods.
	
	\section{System Model and Problem Formulation}
	\label{sec:system_model}
	
	We consider a vehicular network segment, where $\mathcal{N} = \{1,2,\ldots,N\}$ denotes the set of vehicles, with $N$ varying over time as vehicles enter and leave the communication range. The system operates with updates every second, selecting the middle frame of each second as the representative state. Each vehicle is equipped with V2V communication capabilities operating at 5.9 GHz, and the maximum number of hops for any communication path is constrained by $H$. Channel conditions follow urban environment characteristics with path loss exponent $\gamma$.
	
	\subsection{Definition of Decision Variables}	
	For each vehicle \( i \in \mathcal{N} \), we define the decision variables as follows:
	\[
	\mathbf{x}_i = \left[S_b^i, P_n^i, \mathbf{r}_i \right]^T,
	\]
	where:
	\begin{itemize}
		\item \( S_b^i \in [10^5, 10^7] \) represents the data block size,
		\item \( P_n^i \in [10^3, 10^5] \) denotes the node power,
		\item \( \mathbf{r}_i = [r_{i,1}, \dots, r_{i,H}] \) is the relay selection vector, with each element \( r_{i,h} \) representing the relay node at hop \( h \), and \( r_{i,h} \in \mathcal{N}, \, h = 1, \dots, H \).
	\end{itemize}
	
	The complete decision variable vector for all vehicles is given by:
	\[
	\mathbf{x} = \left[\mathbf{x}_1^T, \mathbf{x}_2^T, \dots, \mathbf{x}_N^T \right]^T \in \mathbb{R}^{N(2 + H)}.
	\]
	
	The decision variables must satisfy the following constraint:
	\[
	d(i, r_{i,h}) \leq d_{\text{max}}, \quad \forall i \in \mathcal{N}, h \in \{1, \dots, H\},
	\]
	where \( d(i, r_{i,h}) \) represents the distance between vehicle \( i \) and relay node \( r_{i,h} \), and \( d_{\text{max}} \) is the maximum allowable distance.

	\subsection{Optimization Objectives and Their Significance}
	The proposed framework optimizes the following four critical objectives, each addressing a key aspect of vehicular network performance:
	
	\subsubsection{Objective 1: Communication Latency}
	Minimizing communication delay is essential for supporting real-time applications such as collision avoidance and autonomous driving. The end-to-end delay for vehicle $i$ along path $\mathcal{P}_i$ is expressed as:
	\begin{equation}
		D_i = \sum_{h=1}^{H} \left(\frac{S_{i,j_h}}{B_{i,j_h}} + \delta_{i,j_h}\right),
	\end{equation}
	where $S_{i,j_h}$ is the data size, $B_{i,j_h}$ is allocated bandwidth, and $\delta_{i,j_h}$ accounts for propagation delay. The average network delay is minimized as:
	\begin{equation}
		f_1 = \frac{1}{N}\sum_{i=1}^{N} D_i.
	\end{equation}
	This objective is constrained by SINR requirements and maximum allowable delay, as detailed in Section \ref{sec:constraints}.
	
	\subsubsection{Objective 2: Load Distribution}
	We define the normalized relay load vector as \( \mathbf{\hat{L}} = [\hat{L}_1, \hat{L}_2, \ldots, \hat{L}_N]^\mathrm{T} \in \mathbb{R}^N \), where \( \hat{L}_i \) represents the normalized load of vehicle \( i \):
	\begin{equation}
		\hat{L}_i = \frac{\sum_{j \in \mathcal{N}} \sum_{h=1}^H \mathbbm{1}_{ij}^h}{\sum_{j \in \mathcal{N}} \sum_{h=1}^H 1},
	\end{equation}
	where \( \mathbbm{1}_{ij}^h \) is the indicator function:
	\begin{equation}
		\mathbbm{1}_{ij}^h = 
		\begin{cases}
			1, & \text{if } r_{j,h} = i, \\
			0, & \text{otherwise}.
		\end{cases}
	\end{equation}
	The load distribution objective is then formulated as:
	\begin{equation}
		f_2 = \text{Var}(\mathbf{\hat{L}}) = \frac{1}{N} \sum_{i=1}^N (\hat{L}_i - \mu_L)^2,
	\end{equation}
	where \( \mu_L = \frac{1}{N} \sum_{i=1}^N \hat{L}_i \) is the mean load across all vehicles.
	
	\subsubsection{Objective 3: Link Quality}
	Let \(\text{SINR}_{i,j}\) denote the Signal-to-Interference-plus-Noise Ratio (SINR) between vehicles \(i\) and \(j\):
	\begin{equation}
		\text{SINR}_{i,j} = \frac{P_{rx}(i,j)}{\sum_{k \in \mathcal{N} \setminus \{i,j\}} P_{rx}(k,j) + N_0},
	\end{equation}
	where \(N_0 = k_B T B\) represents the thermal noise power, with \(k_B\) being the Boltzmann constant, \(T\) the temperature in Kelvin, and \(B\) the bandwidth. The received power \(P_{rx}(i,j)\) from vehicle \(i\) to vehicle \(j\) is given by:
	\begin{equation}
		P_{rx}(i,j) = P_t \cdot \text{PL}(d_{i,j}) \cdot \xi_s \cdot G_a \cdot \eta_{\text{Doppler}} / L_s,
	\end{equation}
	where \(P_t\) is the transmission power, \(\xi_s\) is the shadow fading factor, \(G_a\) is the antenna gain, and \(L_s\) is the system loss factor. The path loss \(\text{PL}(d_{i,j})\) is modeled as:
	\begin{equation}
		\begin{split}
			\text{PL}(d_{i,j}) = &\ 20\log_{10}\left(\frac{d_{i,j}}{d_0}\right) + 10\gamma \log_{10}\left(\frac{d_{i,j}}{d_0}\right) \\
			&+ 20\log_{10}\left(\frac{f_c}{10^9}\right),
		\end{split}
	\end{equation}
	where \(d_0\) is the reference distance, \(\gamma\) is the path loss exponent, and \(f_c\) is the carrier frequency. The Doppler effect factor \(\eta_{\text{Doppler}}\) is given by:
	\begin{equation}
		\eta_{\text{Doppler}} = \exp\left(-\min\left(\frac{|\Delta f|}{f_{\text{th}}}, 10\right)\right),
	\end{equation}
	where \(\Delta f\) is the Doppler frequency shift, calculated as \(\Delta f = \frac{v_r f_c}{c}\), with \(v_r\) being the relative velocity, \(c\) the speed of light, and \(f_{\text{th}}\) the threshold frequency. The link quality objective is then formulated as:
	\begin{equation}
		f_3 = \frac{1}{NH}\sum_{i=1}^{N} \sum_{h=1}^{H} \frac{1}{\text{SINR}_{i,r_{i,h}}},
	\end{equation}
	where \(r_{i,h}\) is the selected relay node at hop \(h\) for vehicle \(i\), as defined in the decision variables.

	\subsubsection{Objective 4: Temporal Stability}
	The temporal stability objective ensures consistent communication paths between consecutive time steps. Let \(\mathbf{R}(t)\) denote the relay selection matrix at time \(t\). When the dimensions remain unchanged, the stability metric is defined as:
	\begin{equation}
		f_4 = \frac{\sum_{i=1}^{N} \sum_{h=1}^{H} \mathbbm{1}_{\{r_{i,h}(t) \neq r_{i,h}(t-1)\}}}{NH},
	\end{equation}
	where \(\mathbbm{1}_{\{ \cdot \}}\) is the indicator function, which equals 1 if the condition inside the brackets holds, and 0 otherwise. 
	
	When the dimensions change between time steps, we consider both path changes and dimensional variations. Specifically, we calculate the number of common vehicles and hops between the two time steps, denoted by \(N_c\) and \(H_c\), respectively. The relay selection matrices are compared only for the common vehicles and hops:
	\begin{equation}
		\begin{split}
			f_4 = & \frac{1}{2} \left( 
			\frac{\sum_{i=1}^{N_c} \sum_{h=1}^{H_c} \mathbbm{1}_{\{r_{i,h}(t) \neq r_{i,h}(t-1)\}}}{N_c H_c} \right. \\
			& \left. + \frac{|N(t) - N(t-1)|}{\max(N(t), N(t-1))} 
			\right),
		\end{split}
	\end{equation}
	where:
	\begin{itemize}
		\item \(r_{i,h}(t)\) represents the selected relay node for vehicle \(i\) at hop \(h\) at time \(t\),
		\item \(\mathbbm{1}_{\{ \cdot \}}\) is the indicator function that equals 1 if the condition inside holds and 0 otherwise,
		\item \(N_c\) and \(H_c\) are the common number of vehicles and hops between consecutive time steps. Specifically, \(N_c = \min(N(t), N(t-1))\) and \(H_c = \min(H(t), H(t-1))\), where \(N(t)\) and \(H(t)\) are the number of vehicles and hops at time \(t\),
		\item \(N(t)\) is the number of vehicles at time \(t\).
	\end{itemize}
	The first term considers the path changes for the common vehicles and hops, while the second term penalizes the dimensional variations, reflecting the impact of network size change. The dimensional penalty is calculated as:
	\begin{equation}
		\text{Dimensional Penalty} = \frac{|N(t) - N(t-1)|}{\max(N(t), N(t-1))}.
	\end{equation}
	This objective does not aim to eliminate network dynamics but seeks to mitigate the impact of excessive changes in communication paths and network dimensions. By penalizing significant path variations and large-scale changes in the number of vehicles, the system encourages consistent and reliable communication over consecutive time steps, improving overall network stability and performance.
	
	\subsection{System Constraints}
	\label{sec:constraints}
	The optimization objectives are subject to the following system constraints:
	
	\begin{align}
		P_{rx}(i,j) & \geq P_{th}, \quad \forall (i,j) \in \mathcal{P}, \\
		\text{SINR}_{i,j} & \geq \text{SINR}_{th}, \quad \forall (i,j) \in \mathcal{P}, \\
		D_i & \leq D_{max}, \quad \forall i \in \mathcal{N}, \\
		|\mathcal{P}_i| & \leq H, \quad \forall i \in \mathcal{N}.
	\end{align}
	where:
	\begin{itemize}
		\item \(P_{rx}(i,j)\) is the received power between vehicles \(i\) and \(j\), which must be greater than or equal to the threshold \(P_{th}\) to ensure sufficient signal strength for reliable communication.
		\item \(\text{SINR}_{i,j}\) is the SINR between vehicles \(i\) and \(j\), which must be greater than or equal to the threshold \(\text{SINR}_{th}\) to ensure high-quality communication with minimal interference.
		\item \(D_i\) represents the total delay of vehicle \(i\), which must be less than or equal to the maximum allowable delay \(D_{max}\) to meet the real-time requirements of the network.
		\item \(\mathcal{P}_i\) represents the set of paths associated with vehicle \(i\), and \(H\) is the maximum number of paths or hops a vehicle can be connected to in the network.
	\end{itemize}
	These constraints ensure that the system can find feasible paths that meet the communication quality and delay requirements, while maintaining the stability and reliability of the network.

	\subsection{Optimization Challenges and Contributions}
	The proposed framework faces several significant challenges arising from the dynamic and complex nature of vehicular networks. First, the network's topology is dynamic due to the continuously changing number of vehicles entering and leaving the communication range. This requires adaptive solution representations and flexible encoding of decision variables to maintain robustness and effectiveness as network size fluctuates. Second, the optimization problem involves conflicting objectives, as the framework must balance short-term metrics—such as communication latency (\(f_1\)), load distribution (\(f_2\)), and link quality (\(f_3\))—with long-term temporal stability (\(f_4\)). This trade-off results in a complex optimization landscape, where improving one objective may degrade others, necessitating careful design to achieve a balanced solution. Finally, the real-time nature of vehicular networks imposes strict temporal constraints. With network conditions updating every second, the optimization algorithm must converge rapidly while maintaining high-quality solutions, addressing temporal dependencies, and ensuring smooth adaptation across consecutive time steps. These challenges underscore the need for an advanced optimization approach, motivating the development of the enhanced algorithm presented in the following section.
	
	To address these challenges, this paper introduces an enhanced NSGA-II algorithm with dynamic encoding, elite inheritance, and adaptive constraint handling. These innovations enable efficient and robust optimization in highly dynamic network conditions, ensuring effective trade-offs between conflicting objectives while maintaining real-time performance.
	
	\section{Dynamic Multi-Objective Optimization Algorithm}
	\label{sec:algorithm}
	The optimization of vehicular communication networks demands rapid adaptation to dynamic topology changes while balancing multiple competing objectives: minimizing delay, balancing load, maintaining link quality, and ensuring temporal stability. Real-time constraints require second-level optimization, while temporal continuity across time steps challenges traditional methods that operate independently. These factors drive the design of our enhanced NSGA-II, tailored to address the temporal and dynamic complexities of vehicular networks.
	
	\subsection{Key Innovations in Algorithm}
	\subsubsection{Dynamic Dimension and Genetic Operators Adaptation}
	The dynamic nature of vehicular networks results in varying solution dimensions as vehicles enter and leave the network over time. Let \(N_v(t) = |\mathcal{V}(t)|\) denote the number of vehicles at time \(t\), where \(\mathcal{V}(t)\) represents the set of vehicles in the network. The solution vector \(\mathbf{x}(t)\) encodes decision variables for all vehicles, with a total dimension of \(D(t) = d \cdot N_v(t)\), where \(d\) represents the number of decision variables per vehicle.
	
	To maintain consistency and adaptability, the solution vector \( \mathbf{x}_{\text{new}} \) at time \( t \) is dynamically adjusted based on the vehicle set \( \mathcal{V}(t) \) as follows:
	\begin{equation}
		\mathbf{x}_{\text{new}} = 
		\begin{cases} 
			\text{Truncate}(\mathbf{x}_{\text{old}}, \mathcal{V}_{\text{remain}}), & \text{if } N_v(t) < N_v(t-1), \\
			[\mathbf{x}_{\text{old}}, \mathbf{x}_{\text{gen}}(\mathcal{V}_{\text{new}})], & \text{if } N_v(t) > N_v(t-1),
		\end{cases}
	\end{equation}
	where:
	\begin{itemize}
		\item \( \mathbf{x}_{\text{old}} \) is the solution vector from the previous time step \( t-1 \),
		\item \( \mathcal{V}_{\text{remain}} = \mathcal{V}(t-1) \cap \mathcal{V}(t) \) represents vehicles that remain in the network,
		\item \( \mathcal{V}_{\text{new}} = \mathcal{V}(t) \setminus \mathcal{V}(t-1) \) represents newly added vehicles,
		\item \( \text{Truncate}(\mathbf{x}_{\text{old}}, \mathcal{V}_{\text{remain}}) \) retains decision variables for \( \mathcal{V}_{\text{remain}} \),
		\item \( \mathbf{x}_{\text{gen}}(\mathcal{V}_{\text{new}}) \) generates new decision variables for \( \mathcal{V}_{\text{new}} \).
	\end{itemize}	
	
	The crossover operator implements simulated binary crossover as follows:
	\begin{equation}
		c_1, c_2 = \begin{cases}
			0.5[(y_1 + y_2) \mp \beta|y_2 - y_1|], & \text{if } r < 0.5, \\
			y_1, y_2, & \text{otherwise},
		\end{cases}
	\end{equation}
	where \(\beta\) is calculated as:
	\begin{equation}
		\beta = \begin{cases}
			(2r)^{\frac{1}{\eta + 1}}, & \text{if } r \leq 0.5, \\
			(\frac{1}{2(1-r)})^{\frac{1}{\eta + 1}}, & \text{otherwise},
		\end{cases}
	\end{equation}
	with \(y_1, y_2\) being the parent values, \(r\) is a random number, \(\eta\) is a distribution parameter, and the resulting offspring \(c_1, c_2\) bounded within the feasible range.
	
	The mutation operator adapts its rate based on the magnitude of topology changes:
	\begin{equation}
		p_m^{\text{adaptive}} = p_m \cdot \left(1 + \frac{|\Delta N|}{N_v(t)}\right),
	\end{equation}
	where \(\Delta N = |N_v(t) - N_v(t-1)|\) represents the change in vehicle count between consecutive time steps. This design enhances exploration during significant topology changes while maintaining stability during minimal variations.
	
	\subsubsection{Inheritance Solution Selection}
	The inheritance solution selection follows a two-stage process based on solution quality and diversity. First, solutions from the previous time step are ranked using non-dominated sorting to identify Pareto fronts. Within each front, solutions are further evaluated using a normalized crowding distance metric to maintain diversity:
	\begin{equation}
		d_i = \sum_{m=1}^M \frac{f_i^m(x_{i+1}) - f_i^m(x_{i-1})}{f_{max}^m - f_{min}^m}
		\label{eq:crowding}
	\end{equation}
	where \(d_i\) is the crowding distance of solution \(i\), \(f_i^m\) represents the \(m\)-th objective value, and normalization ensures balanced consideration across objectives. The top ranking solutions are selected based on their non-domination rank and crowding distance.
	
	The inheritance solutions selection is implemented by the pseudo-code in Algorithm \ref{alg:eliteselection}.
	
	\begin{algorithm}[!htpb]
		\caption{Inheritance Solution Selection Process}
		\begin{algorithmic}[1]
			\Require Solutions from previous time step $\text{Pop}(t-1)$
			\Ensure Inheritance solutions $\text{Pop}^*(t-1)$
			\State /* Stage 1: Quality-based Sorting */
			\State Rank solutions using non-dominated sorting into fronts $F_1, F_2, ..., F_k$
			\State /* Stage 2: Diversity Preservation */
			\For{each front $F_i$}
			\For{each objective $m = 1$ to $M$}
			\State Sort solutions in $F_i$ by objective $m$
			\For{each solution $j$ in $F_i$}
			\State Calculate $d_j$ using Equation~\eqref{eq:crowding}
			\EndFor
			\EndFor
			\EndFor
			\State Select solutions based on non-domination rank and crowding distance
			\State \Return Selected inheritance solutions $\text{Pop}^*(t-1)$
		\end{algorithmic}
		\label{alg:eliteselection}
	\end{algorithm}
	
	This two-stage solution selection ensures adaptive optimization under dynamic conditions by maintaining solution validity, promoting inheritance, and enabling efficient adaptation to network changes.
	
	\subsection{Constraint Handling}
	The algorithm implements the constraints through a combination of direct enforcement and optimization objectives. The path length constraint ($|\mathcal{P}_i| \leq H$) is strictly enforced through fixed hop count design. Power and SINR constraints ($P_{rx} \geq P_{th}$, $\text{SINR} \geq \text{SINR}_{th}$) are incorporated into optimization objectives, driving solutions toward better communication quality. Delay constraints ($D_i \leq D_{max}$) are managed through delay-aware path selection in the optimization process.
	
	When vehicle counts change between time steps, solution feasibility is maintained through dimension adjustment: truncating paths when vehicles leave the network, generating new paths for entering vehicles, and preserving valid paths for remaining vehicles. This constraint handling approach effectively ensures solution feasibility while allowing the optimization process to naturally evolve toward better solutions.

	\subsection{Multi-objective Optimization Framework for Vehicular Network in Continuous Scenarios}
	Our enhanced NSGA-II algorithm addresses the dynamic nature of vehicular networks through temporal-aware optimization and adaptive mechanisms. The algorithm incorporates elite inheritance, dimension adaptation, and specialized genetic operators while maintaining the fundamental non-dominated sorting approach of NSGA-II.
	
	The core framework follows a second-by-second optimization approach, with each time step $t$ involving population initialization from previous solutions, iterative evolution through modified genetic operators, and solution selection based on both objectives and diversity. The initialization phase implements a controlled inheritance mechanism where a proportion $\gamma$ of high-performing solutions from the previous time step are preserved and adapted to current network conditions:
	\begin{equation}
		\begin{split}
			\text{Pop}(t) = \{ & \gamma \cdot \text{Pop}^*(t-1) \cup (1-\gamma) \cdot \text{Pop}_{\text{new}}(t)\}
		\end{split}
		\label{eq:inheritance}
	\end{equation}
	where $\text{Pop}^*$ represents elite solutions from time step $t-1$, and $\text{Pop}_{\text{new}}$ contains newly initialized solutions for time step $t$. In	Algorithm \ref{alg:enhanced_nsga2}, we present the complete enhanced NSGA-II framework. The initialization function combines inherited solutions with new random solutions while handling dimension changes due to varying vehicle counts. The genetic operations implement simulated binary crossover (SBX) and adaptive mutation rate based on topology changes. Solution evaluation considers four objectives: communication delay, load balancing, SINR-based link quality, and temporal stability. The selection process combines non-dominated sorting with crowding distance calculation to maintain both convergence and diversity.
	
	\begin{algorithm}[!htpb]
		\caption{Enhanced NSGA-II with Temporal Awareness}
		\label{alg:enhanced_nsga2}
		\begin{algorithmic}[1]
			\Require
			\State Current network state at time $t$
			\State Previous solutions $\text{Pop}^*(t-1)$
			\State Parameters: inheritance ratio $\gamma$
			\Ensure Pareto-optimal routing solutions
			
			\Function{InitializePopulation}{$t$}
			\If{$t == 1$}
			\State Generate random solutions according to $N(1)$
			\Else
			\State Generate solutions based on \eqref{eq:inheritance}
			\EndIf
			\State \Return Combined population
			\EndFunction
			
			\State // Main Algorithm
			\State Update network topology and vehicle count $N(t)$
			\State $population \gets$ \Call{InitializePopulation}{$t$}
			
			\For{$gen \gets 1$ to $max\_generations$}
			\State // Genetic operations
			\State $offspring \gets$ \Call{SBX Crossover}{$population$}
			\State Apply \Call{Adaptive Mutation}{$offspring$}
			
			\State // Evaluate and select
			\For{each solution $s$ in $population \cup offspring$}
			\State \Call{EvaluateObjectives}{$s$}
			\EndFor
			\State Select next generation based on Pareto fronts
			\EndFor
			
			\State \Return First non-dominated front
		\end{algorithmic}
	\end{algorithm}
	
	This enhanced framework effectively balances the need for rapid adaptation to network changes with the maintenance of solution quality and diversity. The temporal inheritance mechanism ensures continuity across time steps, while the adaptive genetic operators enable efficient exploration of the dynamic solution space. The algorithm features a dynamic population size that scales with vehicle density while maintaining computational efficiency. It automatically adjusts the population structure to accommodate varying decision variables based on the current network topology. Solution inheritance is achieved through non-dominated sorting and crowding distance metrics, with simulated binary crossover and polynomial mutation operators facilitating effective exploration of the solution space.

	\subsection{Computational Complexity Analysis}	
	The computational complexity of the proposed algorithm is dominated by the operations involved in the genetic algorithm, including population initialization, crossover, mutation, and evaluation. Let \( N(t) \) represent the number of vehicles in the network at time \( t \), and let \( D(t) \) denote the number of decision variables, where \( D(t) = d \cdot N(t) \), with \( d \) being the number of decision variables per vehicle. The complexity of the population initialization step is \( \mathcal{O}(D(t)) \), as it requires generating decision variables for all vehicles in the network.
	
	The crossover operator is based on simulated binary crossover (SBX). For each pair of solutions, the crossover operation requires constant time, \( \mathcal{O}(d) \), leading to a total complexity of \( \mathcal{O}(D(t) \cdot N_{\text{pop}}) \) for \( N_{\text{pop}} \) solutions. The mutation operation has an adaptive rate based on topology changes. Each mutation is applied to a solution, requiring \( \mathcal{O}(d) \) operations per solution. Thus, the mutation complexity is \( \mathcal{O}(D(t) \cdot N_{\text{pop}}) \). The solution evaluation step involves assessing multiple objectives for each solution. If there are \( M \) objectives, the evaluation complexity is \( \mathcal{O}(M) \), leading to an overall evaluation complexity of \( \mathcal{O}(M \cdot N_{\text{pop}}) \) for \( N_{\text{pop}} \) solutions. The non-dominated sorting process, which is central to the selection mechanism, has a complexity of \( \mathcal{O}(N_{\text{pop}} \cdot \log N_{\text{pop}}) \). This step is followed by crowding distance calculation, which requires \( \mathcal{O}(N_{\text{pop}}) \).
	
	In summary, the overall complexity of the algorithm is dominated by the selection and evaluation steps. The total computational complexity for one generation is given in \eqref{eq:complexity}.	
	\begin{equation}
		\mathcal{O}(D(t) \cdot N_{\text{pop}} + N_{\text{pop}} \cdot \log N_{\text{pop}})
		\label{eq:complexity}
	\end{equation}	
	where \( N_{\text{pop}} \) is the population size and \( M \) is the number of objectives. This complexity analysis shows that the proposed algorithm scales linearly with the number of decision variables and solutions, with a logarithmic dependence on the population size due to the sorting operation.

	\section{Simulation and Discussion}
	\label{sec:sim}
\begin{figure*}[!htbp]
		\centering
		\includegraphics[width=\textwidth]{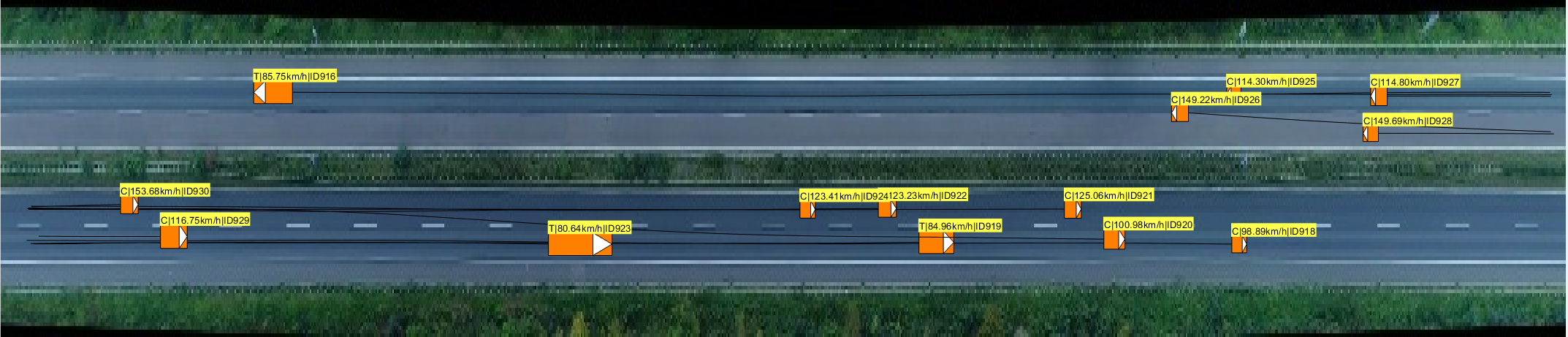} % 图片文件名（无需扩展名）
		\caption{A demo visualization diagram for high-D database} % 添加图片标题
		\label{fig:example} % 添加图片标签，供引用
	\end{figure*}

	In this section, simulations are conducted based on the high-D data-set \cite{highDdataset}, which is a collection of naturalistic vehicle trajectories recorded on German highways using drone technology, providing precise vehicle data with minimal positioning error, overcoming traditional data collection limitations. Based on the data, a visualization diagram is presented as shown in Fig. \ref{fig:example}.
	 
\begin{figure*}[!htpb]
		\centering
		% Subfigure 1
		\begin{subfigure}[b]{0.32\textwidth}
			\centering
			\includegraphics[width=\textwidth]{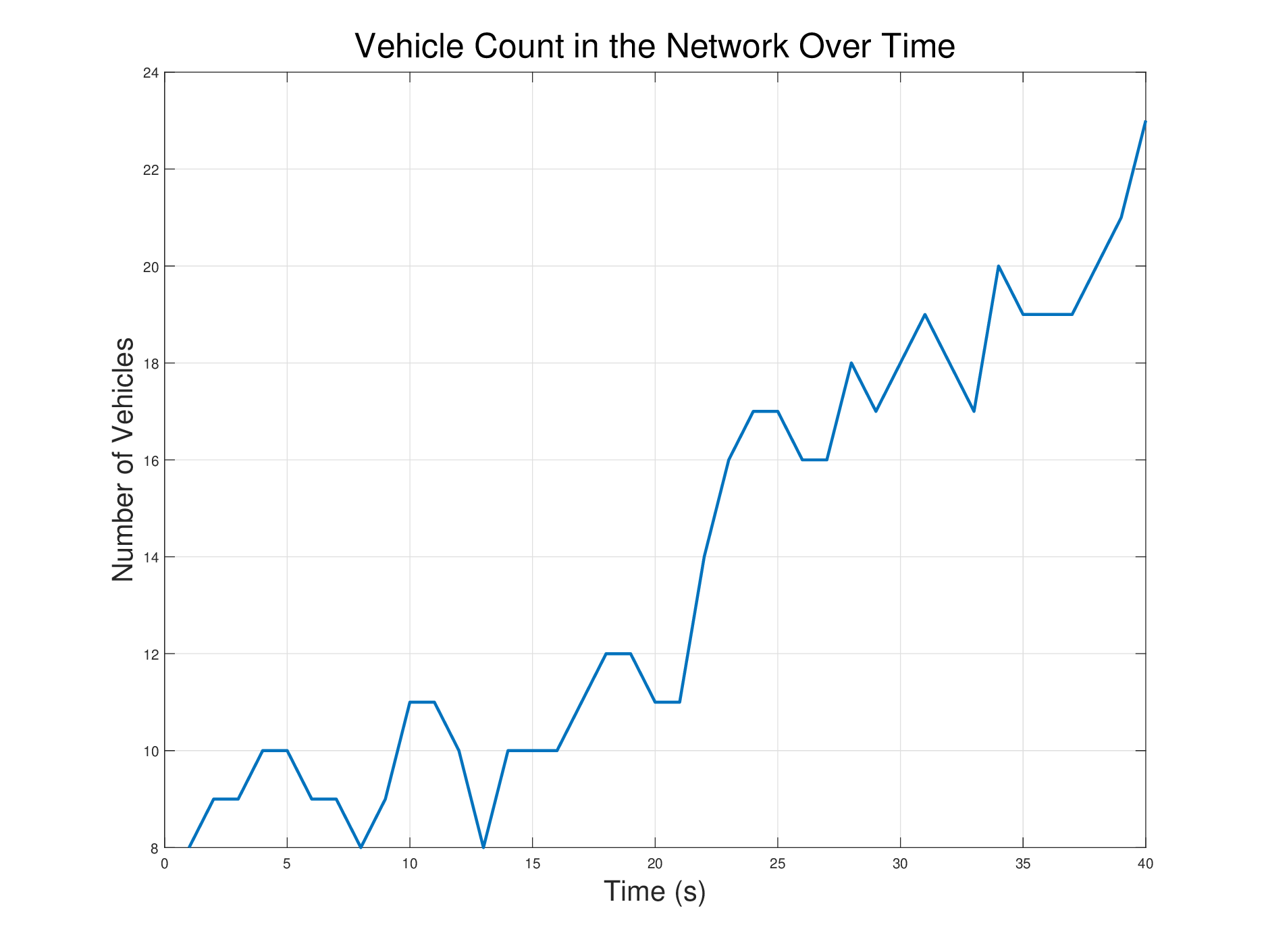}
			\caption{Scenario 1: Gradual increase in the number of vehicles}
			\label{fig:s1}
		\end{subfigure}
		% Subfigure 2
		\begin{subfigure}[b]{0.32\textwidth}
			\centering
			\includegraphics[width=\textwidth]{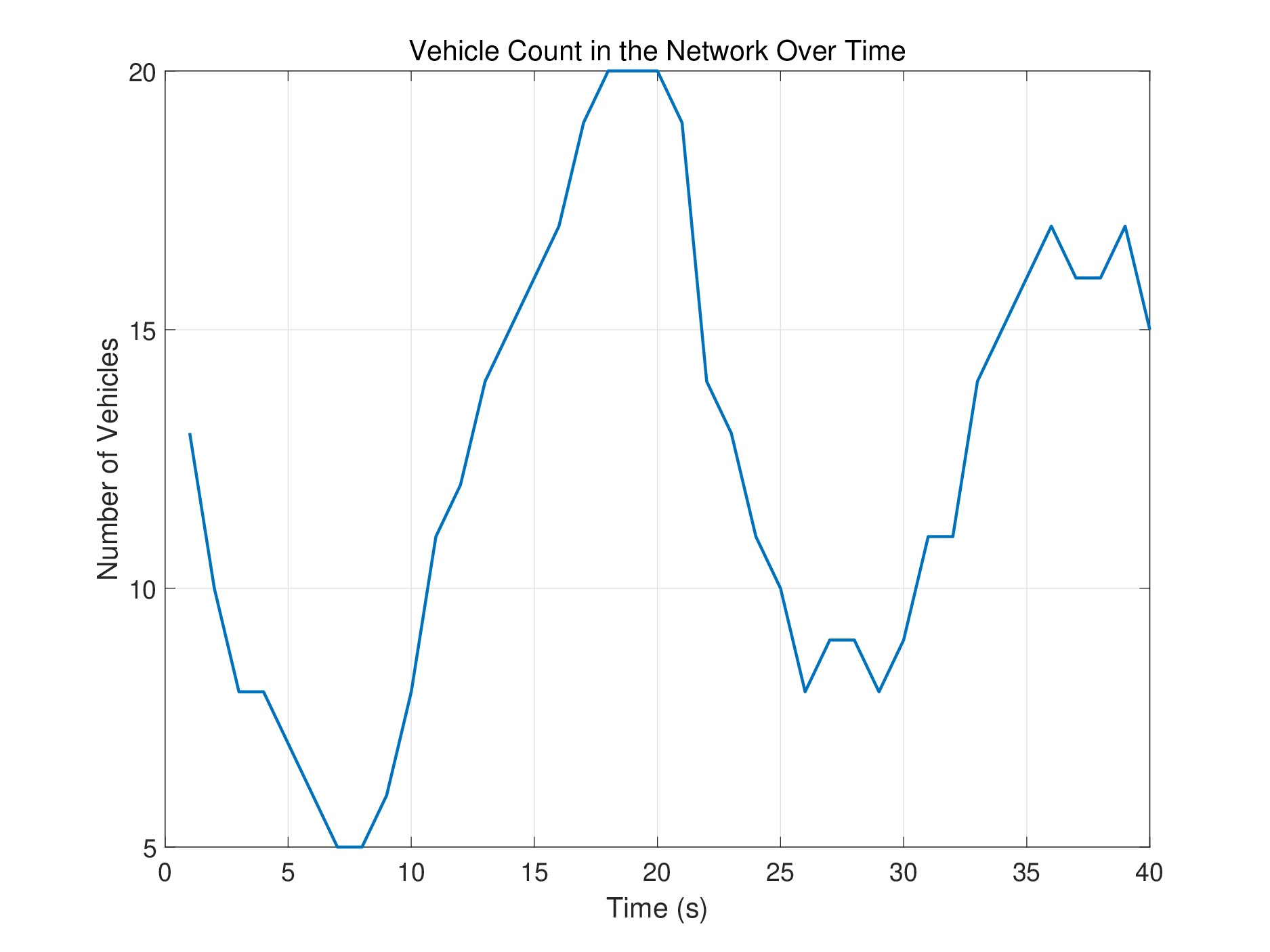}
			\caption{Scenario 2: Fluctuations in the number of vehicles}
			\label{fig:s2}
		\end{subfigure}
		% Subfigure 3
		\begin{subfigure}[b]{0.32\textwidth}
			\centering
			\includegraphics[width=\textwidth]{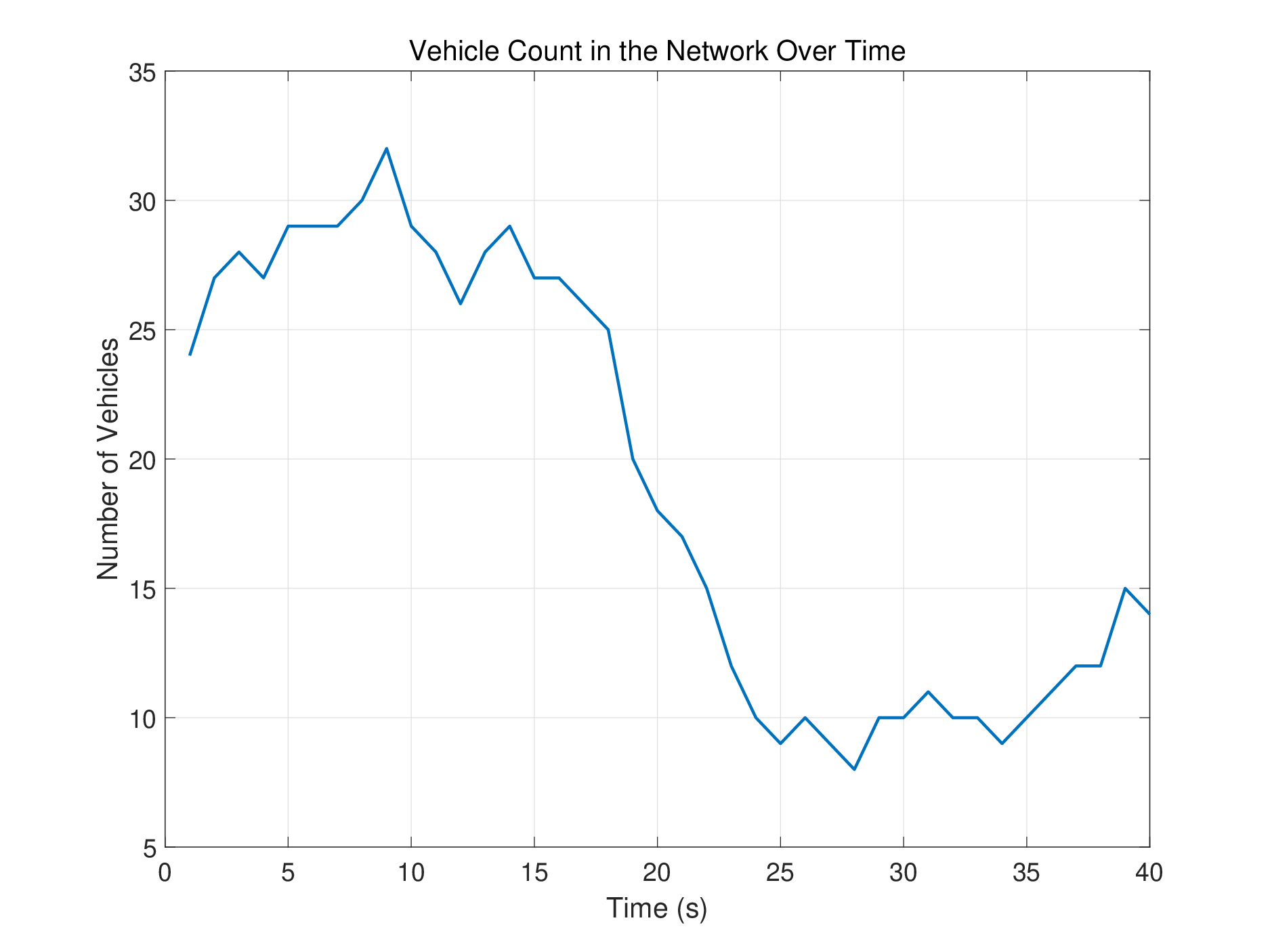}
			\caption{Scenario 3: Gradual decrease in the number of vehicles}
			\label{fig:s3}
		\end{subfigure}
		\caption{Changes in the number of vehicles within the scenarios over 1000 frames (40 seconds)}
		\label{fig:s}
	\end{figure*}

	In this database, we conduct simulation experiments based on highway scenarios. Three different scenarios are selected for the algorithm simulation, each lasting approximately 40 seconds. During these 40 seconds, vehicles in the scenario either exit or enter the scene. The three chosen scenarios represent different traffic dynamics\footnote{The scenarios are derived from the high-D dataset, specifically from data record 1, frames 3000 to 4000; data record 2, frames 1 to 1000; and data record 2, frames 1000 to 2000}: one with a higher number of vehicles entering than exiting, one with fluctuating numbers of vehicles entering and exiting, and one with more vehicles exiting than entering. The vehicle counts are presented in Fig~\ref{fig:s}. Using high-D data, we present the results in the form of a dynamic video to assist readers in understanding the scenarios, which will be uploaded as supplementary material.  
	
	\subsection{Simulation Environment Setting}
	The experimental environment utilizes MATLAB 2024b running on a Windows operating system with an Intel Core i7-13790F CPU and 32GB memory. Each vehicle is associated with three decision variables: communication bandwidth, transmission power, and path. The path can have up to 3 hops, resulting in a decision dimension of 5 per vehicle. To enhance the responsiveness of the algorithm to the dynamic nature of the VANET scenario, we used moderate computational resources by setting the population size to 100 and performing 20 iterations per second. Four experiments were conducted based on different inheritance ratios, where the proportion of individuals inherited from the previous second's scenario was set to 0, 0.3, 0.5, and 0.8, respectively. The simulation parameters are configured as follows in Table \ref{tab:parameters}.
	
	\begin{table*}[!htbp]
		\centering
		\caption{Parameters for System and Algorithm}
		\label{tab:parameters}
		\begin{tabular}{|l|l|}
			\hline
			\textbf{Category}                  & \textbf{Parameters}                                                                                     \\ \hline
			\multirow{3}{*}{Communication System} 
			& Network bandwidth ($B_{net}$): 10 MHz                                                                   \\
			& Transmission power ($P_t$): 10 W                                                                         \\
			& Communication range: 2 m (min), 300 m (max), Reference distance ($d_0$): 1 m                            \\ \hline
			\multirow{2}{*}{Channel and Interference} 
			& Path loss exponent ($\gamma$): 2.5, Antenna gain: 2, System loss: 1                                      \\
			& Shadow fading: 4 dB, Rice factor: 10 dB, Doppler threshold: 1000 Hz, Environment: urban, $T$: 290 K     \\ \hline
			\multirow{2}{*}{Optimization Algorithm} 
			& Population size: 200, Max generations: 10, Crossover rate: 0.9, Mutation rate: 0.1, Tournament size: 2   \\
			& Time continuity weight ($w_c$): 0.3, Inheritance ratio ($\gamma$): 0, 0.3, 0.5 and 0.8                                  \\ \hline
			QoS Parameters                     & Delay threshold: 100 ms, Reliability threshold: 0.9, Min SINR: 10 dB                                    \\ \hline
			Temporal Parameters                & Frame rate: 25 fps, Analysis window: 500 frames (20 s), Update interval: 1 s                            \\ \hline
		\end{tabular}
	\end{table*}

	\subsection{Simulation Results and Analysis}
	The overall experimental design is as follows: simulation experiments are conducted across three distinct scenarios based on the variations in the number of vehicles within the scene (as shown in Fig.~\ref{fig:s}). Each experiment demonstrates the feasibility and effectiveness of the proposed algorithm in addressing the challenges of dynamic multi-objective optimization caused by variations in vehicle density in dynamic environments. Furthermore, leveraging the inherent correlation between the number of frames and seconds in the scenarios, the experiments investigate the impact of inheritance ratios of the population under different temporal conditions on the algorithm's performance. This study provides valuable insights into achieving computational efficiency while maintaining high optimization effectiveness in dynamic vehicular network scenarios. The detailed simulation results and corresponding analyses are presented in the following subsections.
	
	\subsubsection{Scenario 1}
\begin{figure*}[!htbp]
		\centering
		% 第一个子图
		\begin{subfigure}{0.24\textwidth}
			\centering
			\includegraphics[width=\textwidth]{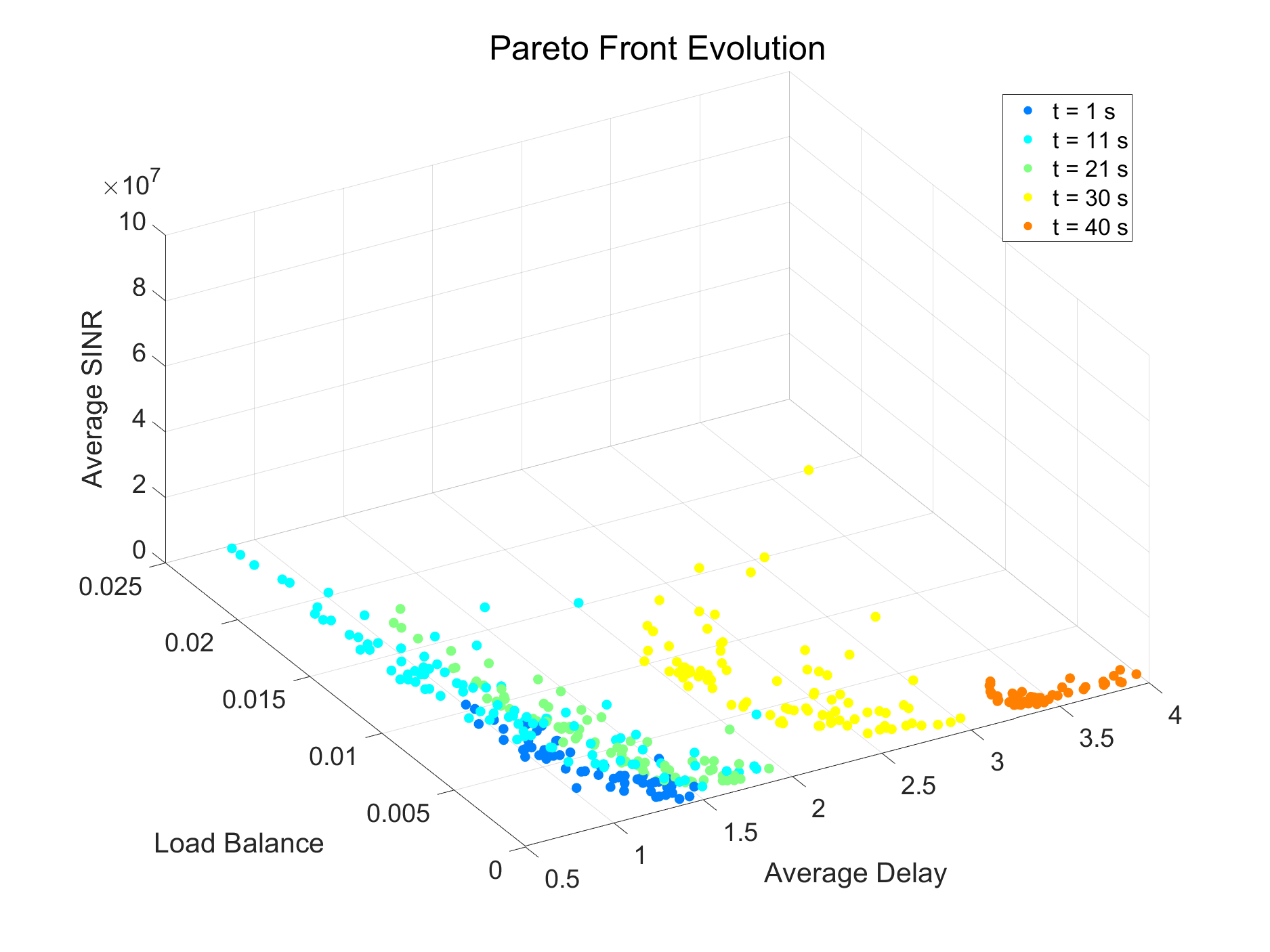}
			\caption{Inheritance Proportion: 0}
			\label{fig:pareto00_s1}
		\end{subfigure}
		% 第二个子图
		\begin{subfigure}{0.24\textwidth}
			\centering
			\includegraphics[width=\textwidth]{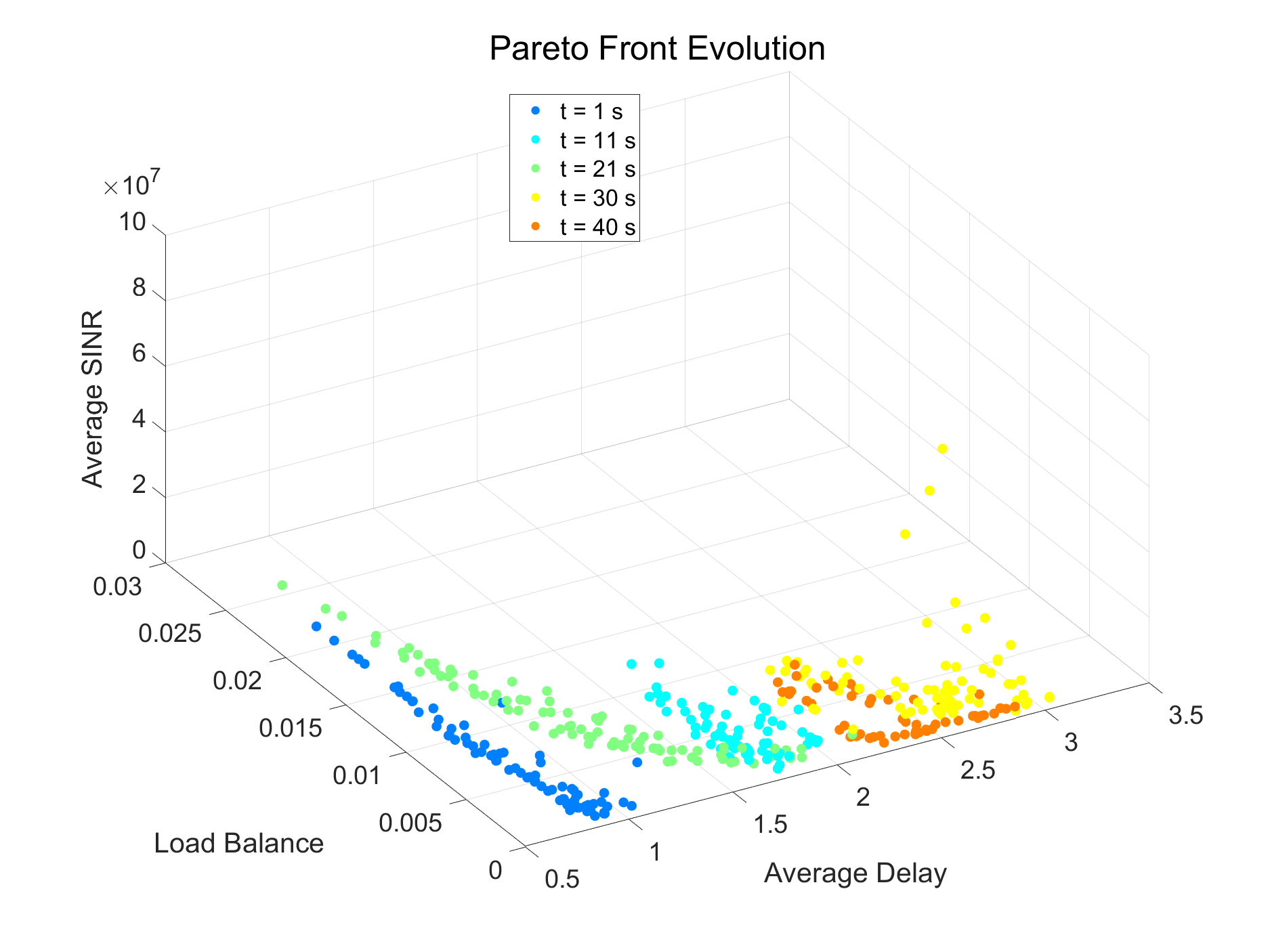}
			\caption{Inheritance Proportion: 30\%}
			\label{fig:pareto03_s1}
		\end{subfigure}
		% 第三个子图
		\begin{subfigure}{0.24\textwidth}
			\centering
			\includegraphics[width=\textwidth]{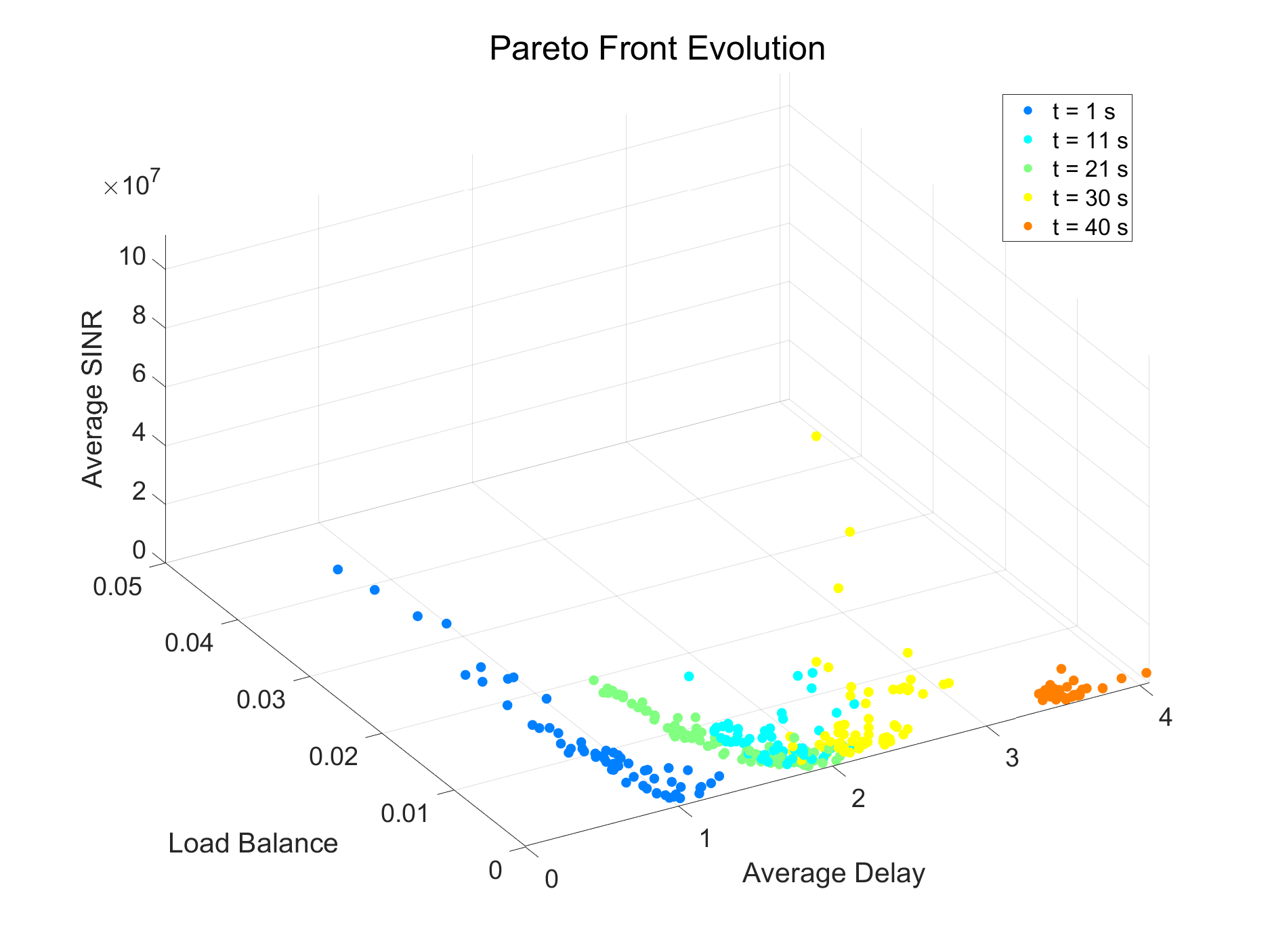}
			\caption{Inheritance Proportion: 50\%}
			\label{fig:pareto05_s1}
		\end{subfigure}
		% 第四个子图
		\begin{subfigure}{0.24\textwidth}
			\centering
			\includegraphics[width=\textwidth]{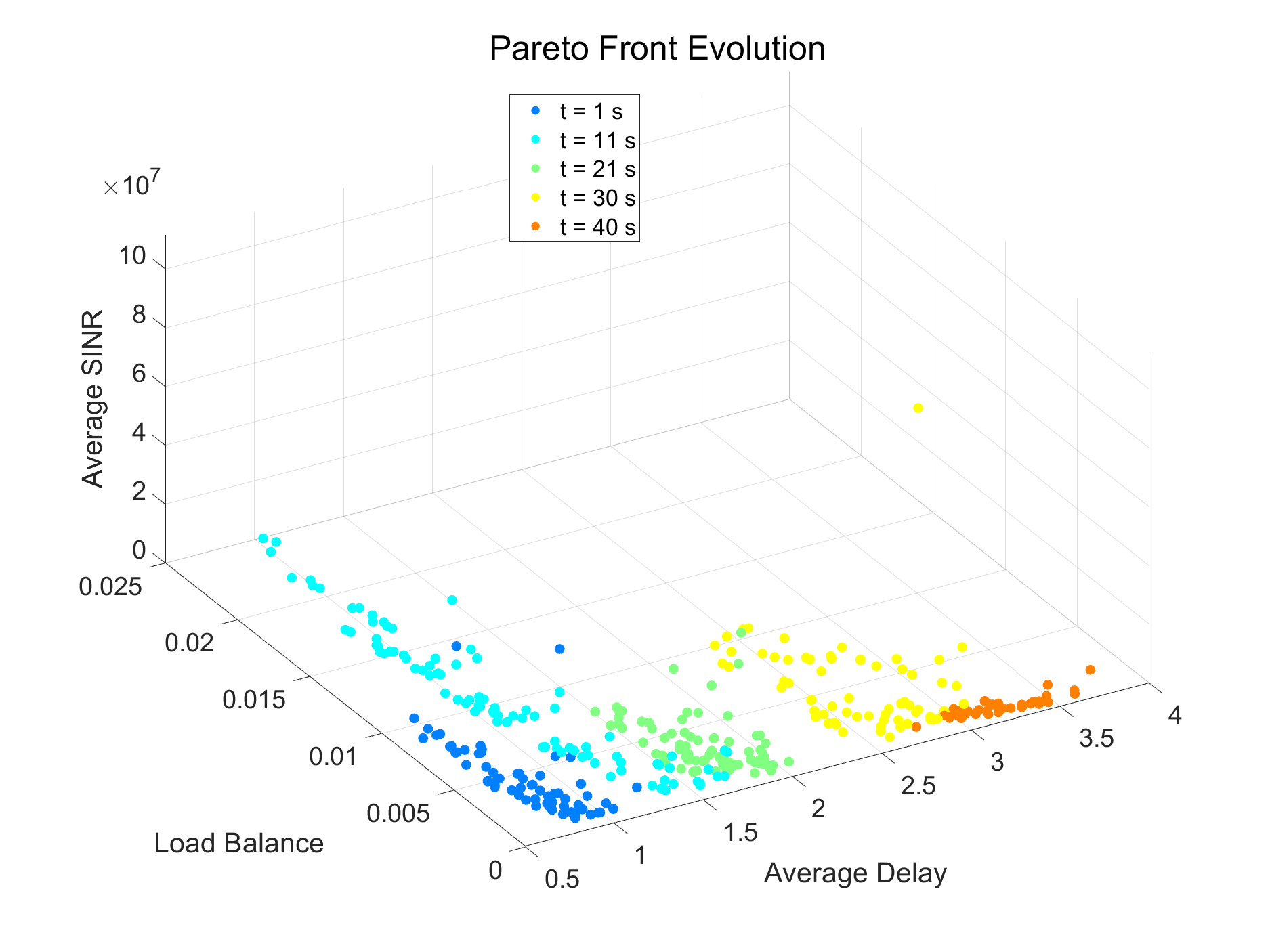}
			\caption{Inheritance Proportion: 80\%}
			\label{fig:pareto08_s1}
		\end{subfigure}
		\caption{Pareto Fronts with Different Inheritance Proportion on Scenario 1}
		\label{fig:pareto_fronts_s1}
	\end{figure*}

	The algorithm's performance in Scenario 1 is illustrated in Fig.~\ref{fig:pareto_fronts_s1}, where the distribution of Pareto solutions is recorded every 10 seconds. In this scenario, the number of vehicles exhibits an increasing trend, as shown in Fig.~\ref{fig:s1}. Over approximately 40 seconds (1000 frames), the number of vehicles within a 480-meter stretch of highway rises from around 10 vehicles to over 20 vehicles. 
	
	Fig.~\ref{fig:pareto_fronts_s1} illustrates the Pareto solutions across five stages from 1 to 40 seconds, represented by different colors. Fig.~\ref{fig:pareto00_s1} demonstrates the algorithm's performance without the inheritance of elite individuals between seconds. From this figure, it can be observed that the proposed multi-objective optimization method effectively achieves desirable values for delay, load balancing, and SINR under varying time conditions. Because Fig.~\ref{fig:pareto00_s1} represents the algorithm without individual inheritance across scenarios, it can be observed that the proposed algorithm achieves relatively uniform Pareto solutions at various seconds, demonstrating its effectiveness. However, heuristic optimization algorithms typically consume significant computational time and resources. In vehicular network systems, vehicles in consecutive scenarios exhibit certain similarities, particularly in second-level scenario transitions where the topological changes are relatively limited. To address this, a proportion of elite individuals from the NSGA-II algorithm in the previous second are inherited into the next second.
	
	As shown in Fig.~\ref{fig:pareto03_s1}, Fig.~\ref{fig:pareto05_s1}, and Fig.~\ref{fig:pareto08_s1}, the inheritance proportions are 30\%, 50\%, and 80\%, respectively. Based on the distribution of the Pareto solution sets, the proposed algorithm achieves well-shaped Pareto fronts under all three inheritance proportions. Regarding the Average Delay metric, the performance is optimal at the 30\% inheritance proportion, with the intermediate Pareto distributions also outperforming those obtained with higher inheritance proportions. This indicates that in such dynamic scenarios, a moderate inheritance proportion, such as 30\%, effectively enhances the initial solutions for the subsequent scenario. As the inheritance proportion increases, the Pareto front deteriorates. For instance, at a 50\% inheritance proportion, the solutions are less diverse, showing a more clustered distribution, and the performance metric, such as Average Delay, starts to degrade. In later stages, the shape of the Pareto front worsens further, and performance metrics, including Average Delay, are adversely affected compared to those at lower inheritance proportions.
	
\begin{figure*}[!htbp]
		\centering
		% 子图1
		\begin{subfigure}[b]{0.24\textwidth}
			\includegraphics[width=\textwidth]{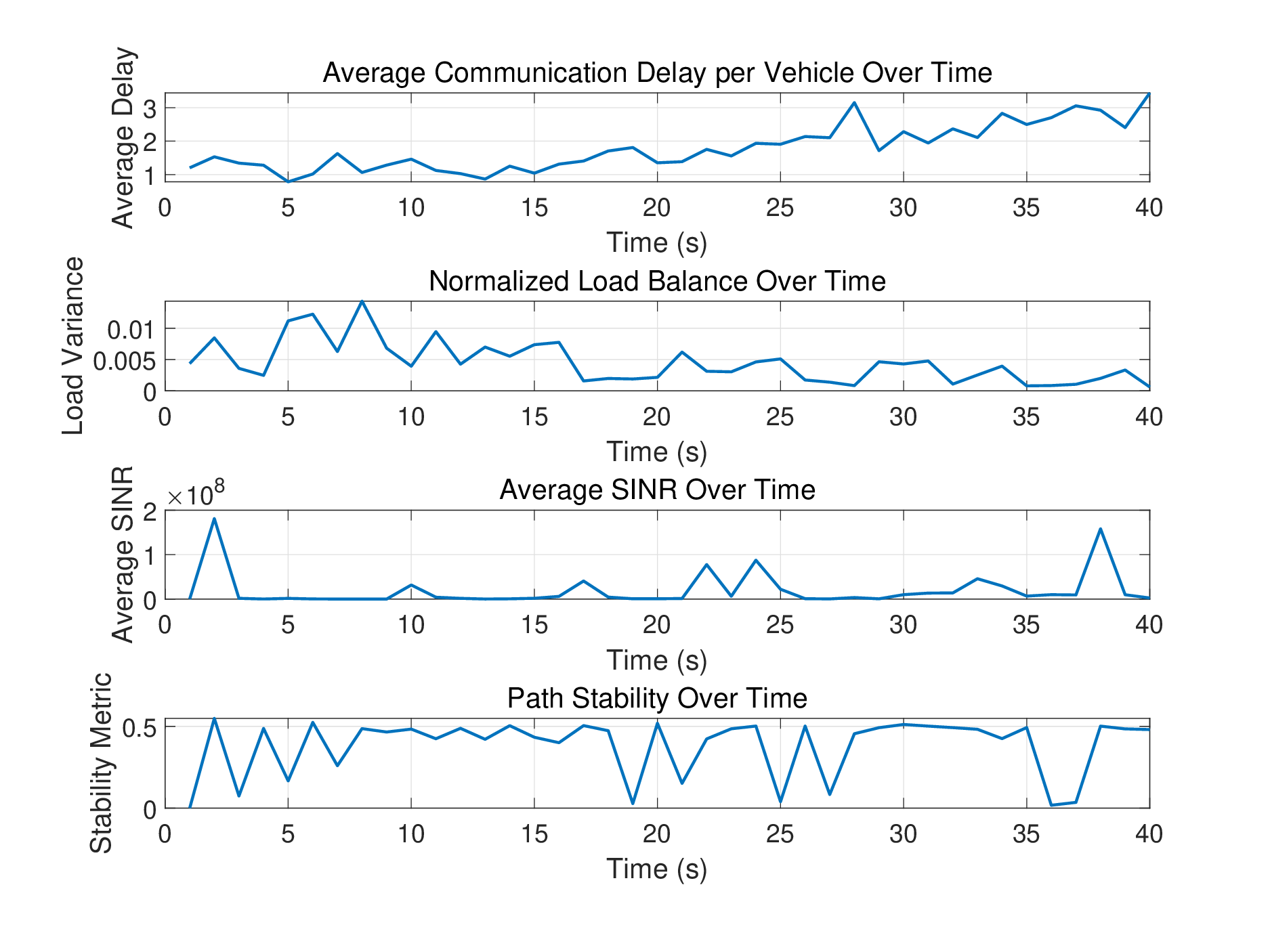}
			\caption{Performance with inheritance ratio: 0.0}
			\label{fig:perf_00_s1}
		\end{subfigure}
		\hfill
		% 子图2
		\begin{subfigure}[b]{0.24\textwidth}
			\includegraphics[width=\textwidth]{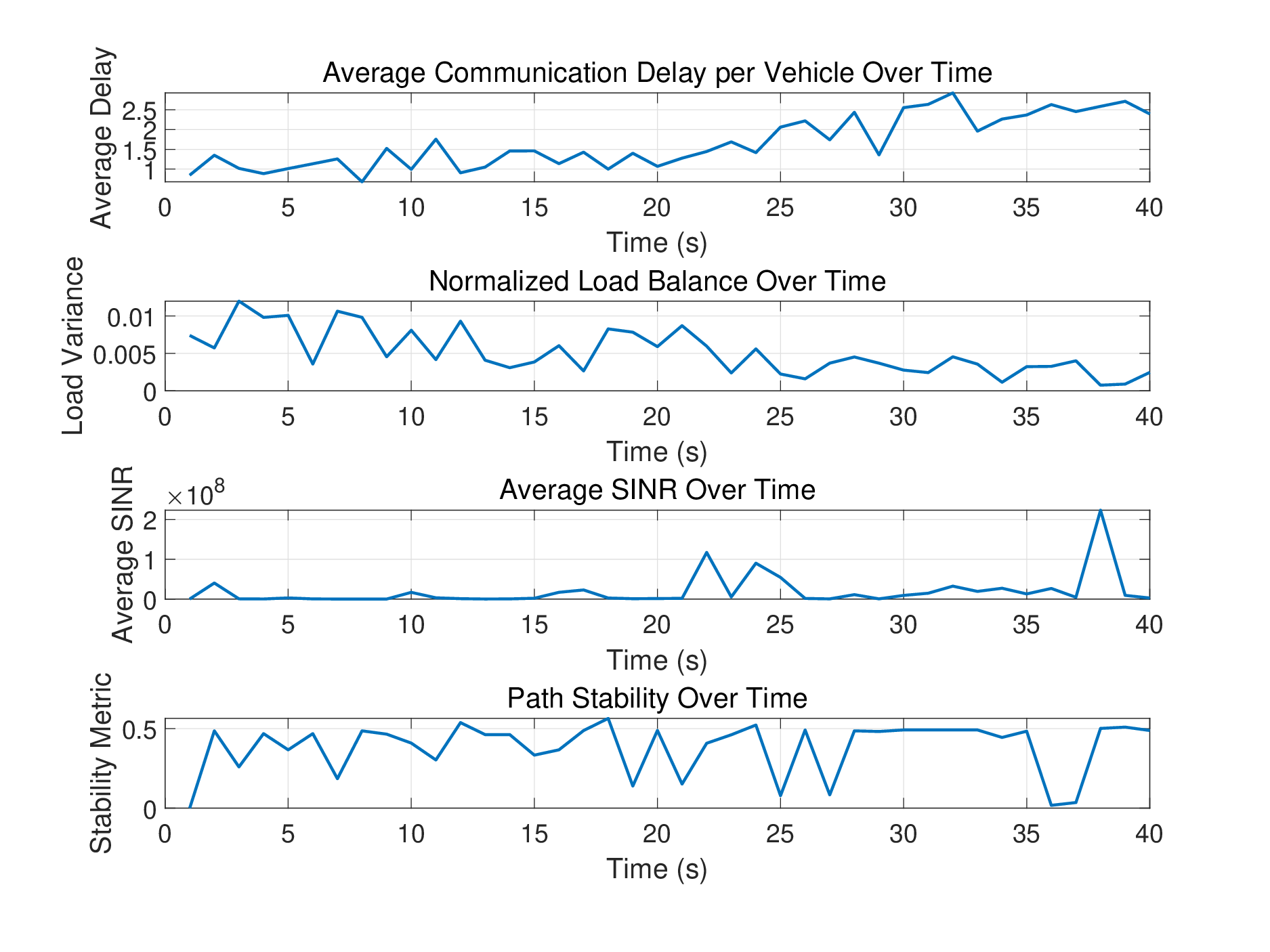}
			\caption{Performance with inheritance ratio: 0.3}
			\label{fig:perf_03_s1}
		\end{subfigure}
		\vspace{0.5cm}
		% 子图3
		\begin{subfigure}[b]{0.24\textwidth}
			\includegraphics[width=\textwidth]{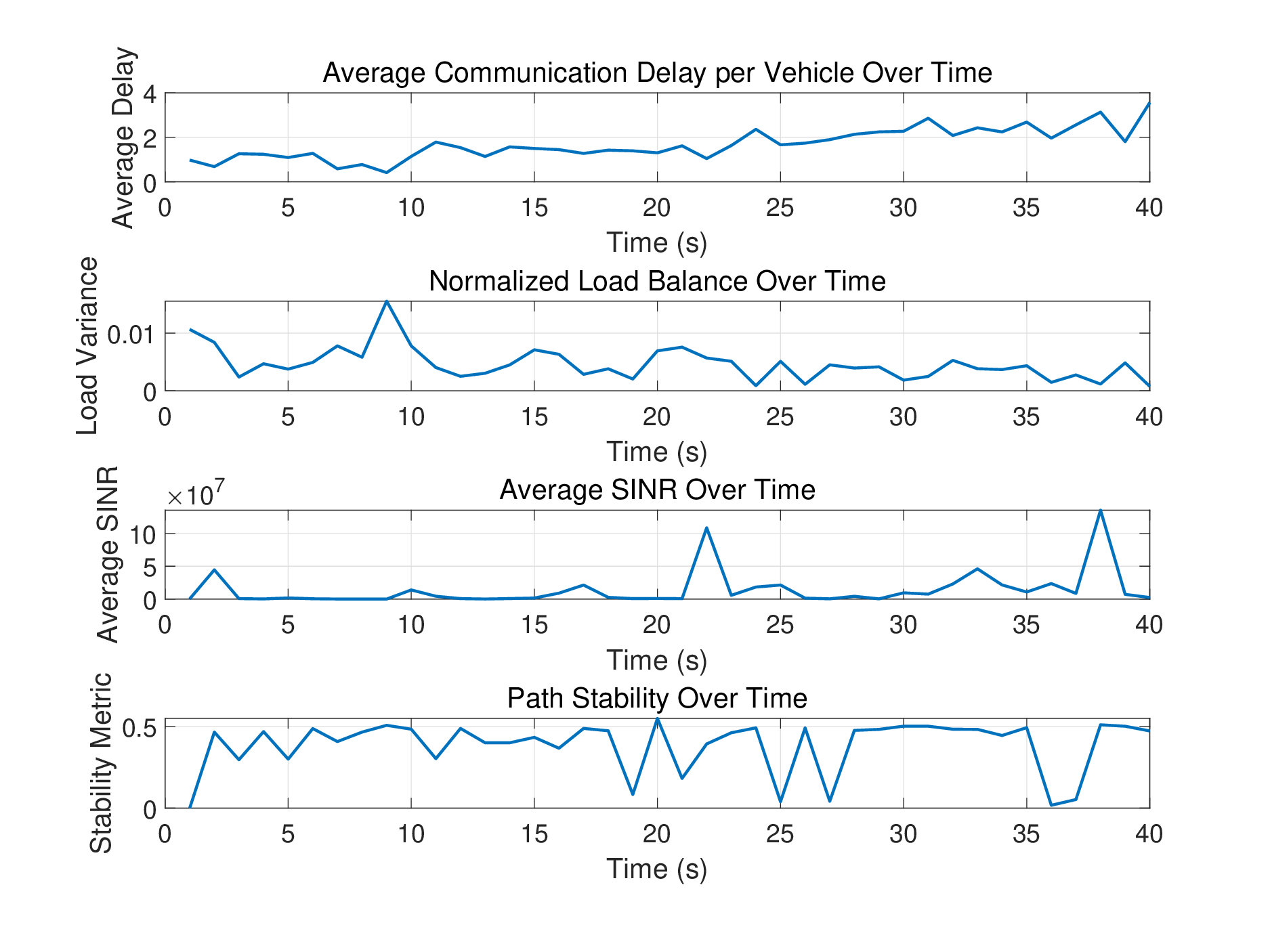}
			\caption{Performance with inheritance ratio: 0.5}
			\label{fig:perf_05_s1}
		\end{subfigure}
		\hfill
		% 子图4
		\begin{subfigure}[b]{0.24\textwidth}
			\includegraphics[width=\textwidth]{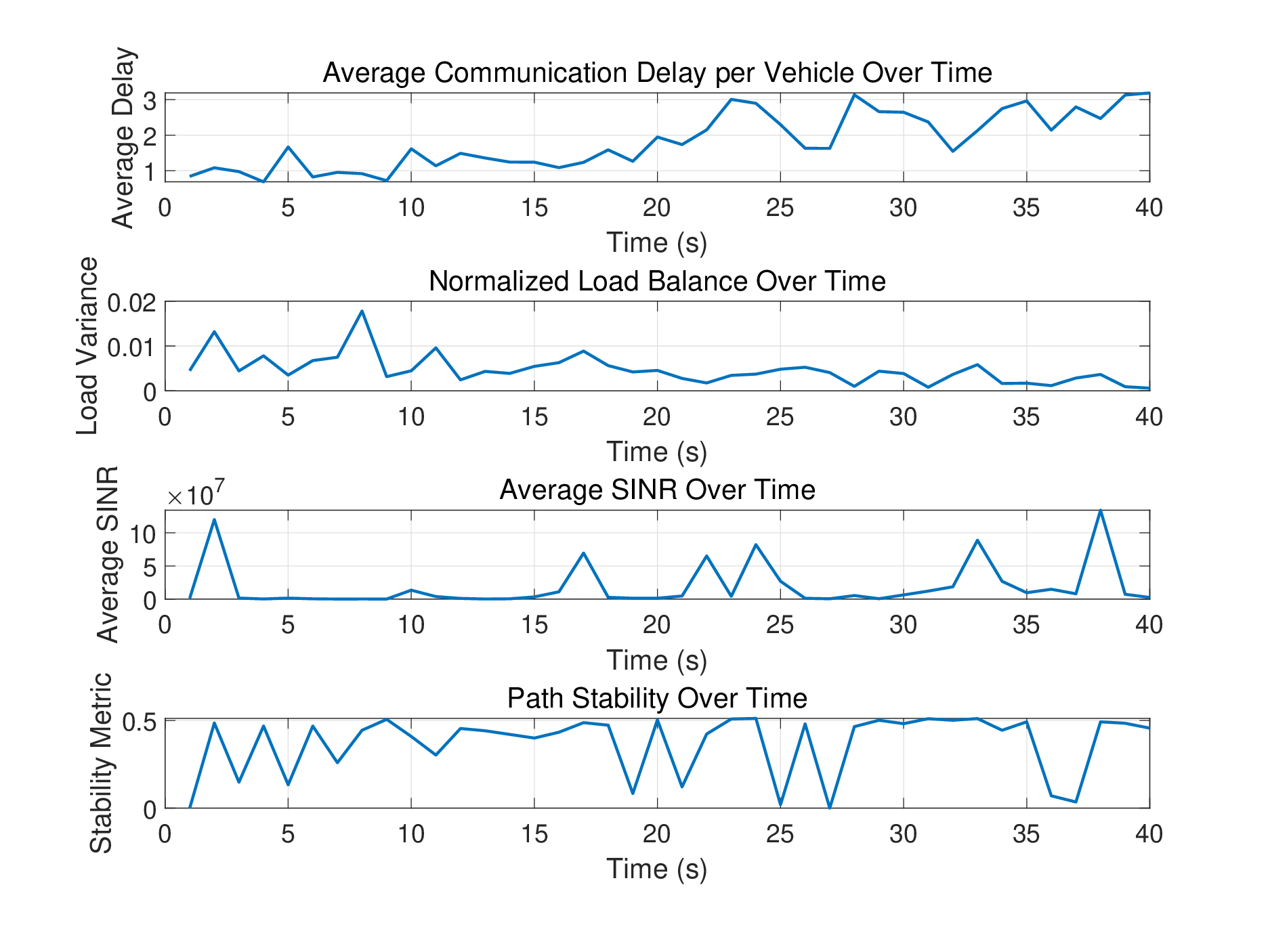}
			\caption{Performance with inheritance ratio: 0.8}
			\label{fig:perf_08_s1}
		\end{subfigure}
		\caption{Performances of Different Inheritance Proportion on Scenario 1}
		\label{fig:performance_s1}
	\end{figure*}

	The performance of the proposed algorithm was evaluated under varying inheritance ratios, as illustrated in Fig.~\ref{fig:performance_s1}. Based on the experimental results, we investigate the impact of different inheritance ratios (0.0, 0.3, 0.5, and 0.8) on algorithm performance through four key metrics: average communication delay, normalized load balance, average SINR, and path stability.
	
	The analysis of average communication delay reveals a clear correlation between delay performance and inheritance ratio. With zero inheritance (0.0), the system maintains a stable delay around 2 units with minimal fluctuations. As the inheritance ratio increases, both the average delay value and its variance show notable deterioration. Particularly, at 0.8 inheritance ratio, the system exhibits the highest delay values and maximum fluctuations, indicating that excessive inheritance may compromise communication efficiency.
	
	The normalized load balance metric demonstrates that all inheritance ratios achieve relatively balanced load distribution, with values consistently remaining around 0.01. However, subtle differences can be observed. The 0.0 and 0.3 inheritance cases present the more stable load distribution with minimal variations, while higher inheritance ratios, especially 0.8, introduce more pronounced fluctuations reaching up to 0.02. This suggests that increased inheritance slightly compromises the system's ability to maintain uniform load distribution.
	
	Regarding SINR performance, all cases exhibit periodic fluctuation patterns, but with varying intensities. Lower inheritance ratios (0.0 and 0.3) maintain relatively stable SINR variations, whereas higher ratios (0.5 and 0.8) lead to more significant fluctuations, particularly in later time periods. This phenomenon suggests that excessive inheritance may adversely affect the system's signal quality stability.
	
	Path stability analysis reveals an interesting trade-off. Zero inheritance results in the highest path variability due to the absence of historical solution influence. As the inheritance ratio increases, path stability improves consistently, with 0.8 inheritance ratio achieving the best stability despite minor fluctuations. This observation confirms the positive correlation between inheritance ratio and path stability.
	
	Comprehensive evaluation of these metrics suggests that a moderate inheritance ratio of 0.3 represents an optimal balance. This configuration maintains satisfactory performance in delay, load balance, and SINR metrics while achieving improved path stability compared to zero inheritance. Higher inheritance ratios, while beneficial for path stability, lead to significant degradation in other performance metrics. Therefore, the 0.3 inheritance ratio emerges as the most balanced choice for overall system performance, and a moderate inheritance ratio also helps reduce computational load.
	
	\subsubsection{Scenario 2}
\begin{figure*}[!htbp]
		\centering
		% 第一个子图
		\begin{subfigure}{0.24\textwidth}
			\centering
			\includegraphics[width=\textwidth]{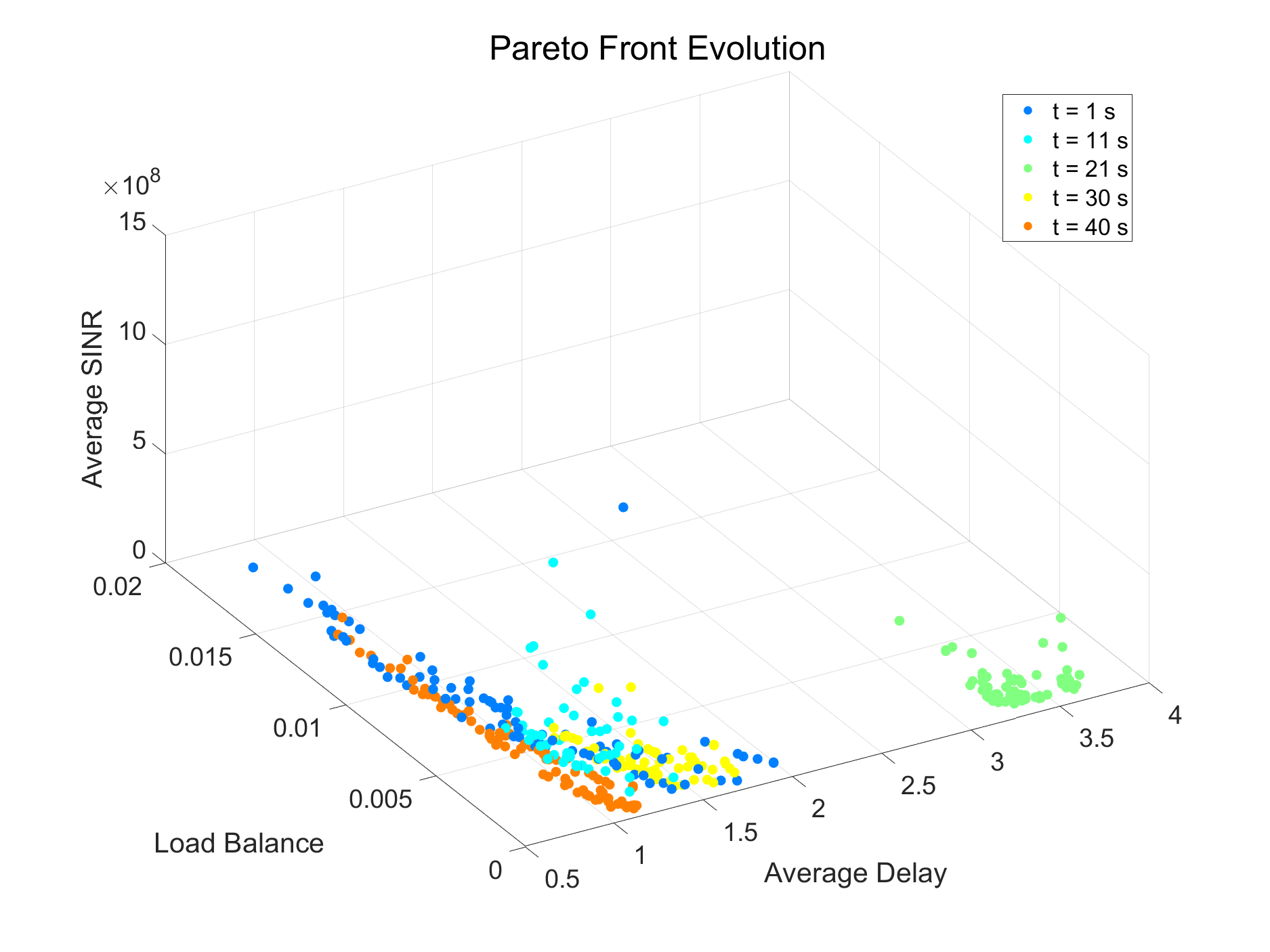}
			\caption{Inheritance Proportion: 0}
			\label{fig:pareto00_s2}
		\end{subfigure}
		% 第二个子图
		\begin{subfigure}{0.24\textwidth}
			\centering
			\includegraphics[width=\textwidth]{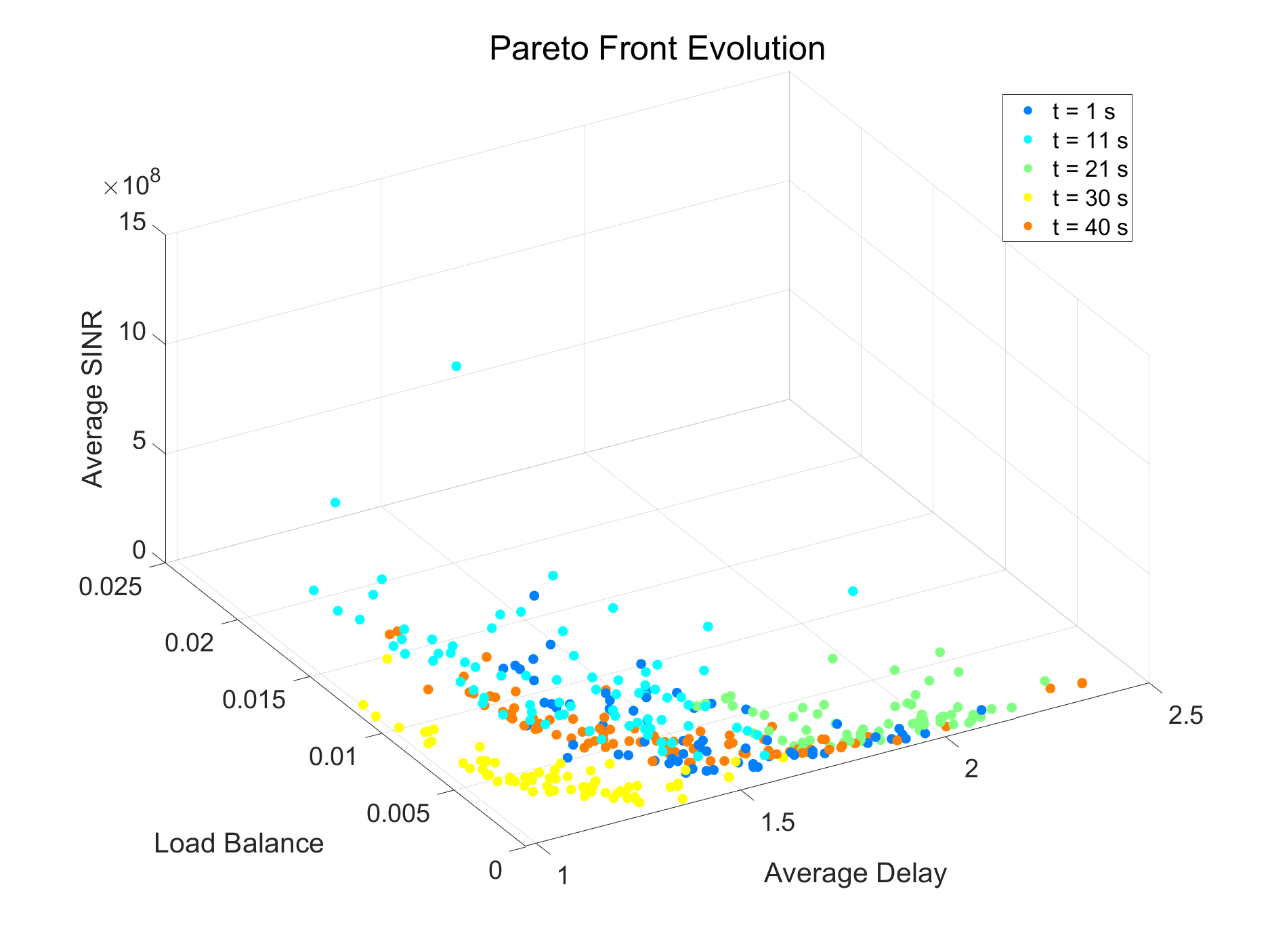}
			\caption{Inheritance Proportion: 30\%}
			\label{fig:pareto03_s2}
		\end{subfigure}
		% 第三个子图
		\begin{subfigure}{0.24\textwidth}
			\centering
			\includegraphics[width=\textwidth]{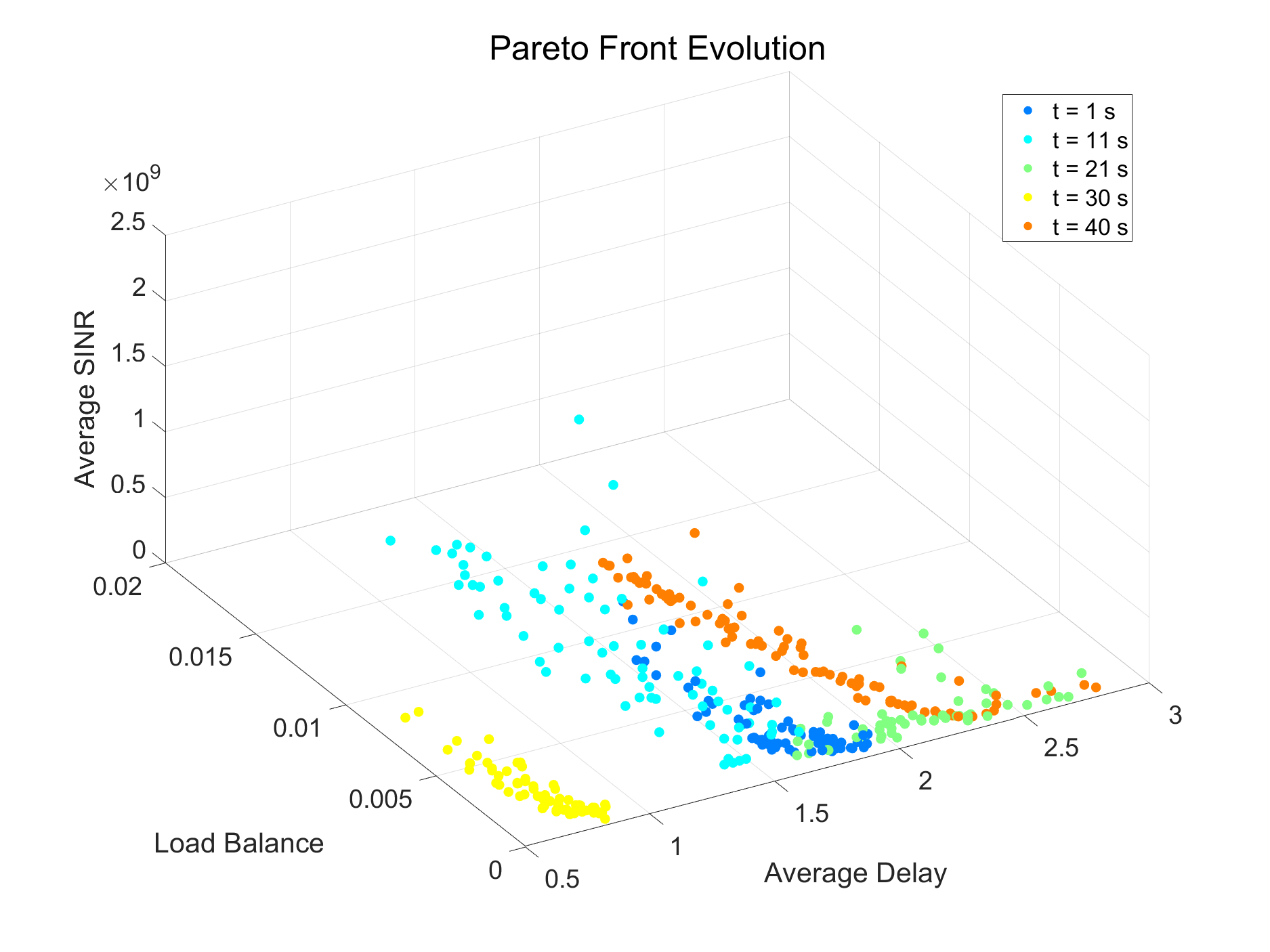}
			\caption{Inheritance Proportion: 50\%}
			\label{fig:pareto05_s2}
		\end{subfigure}
		% 第四个子图
		\begin{subfigure}{0.24\textwidth}
			\centering
			\includegraphics[width=\textwidth]{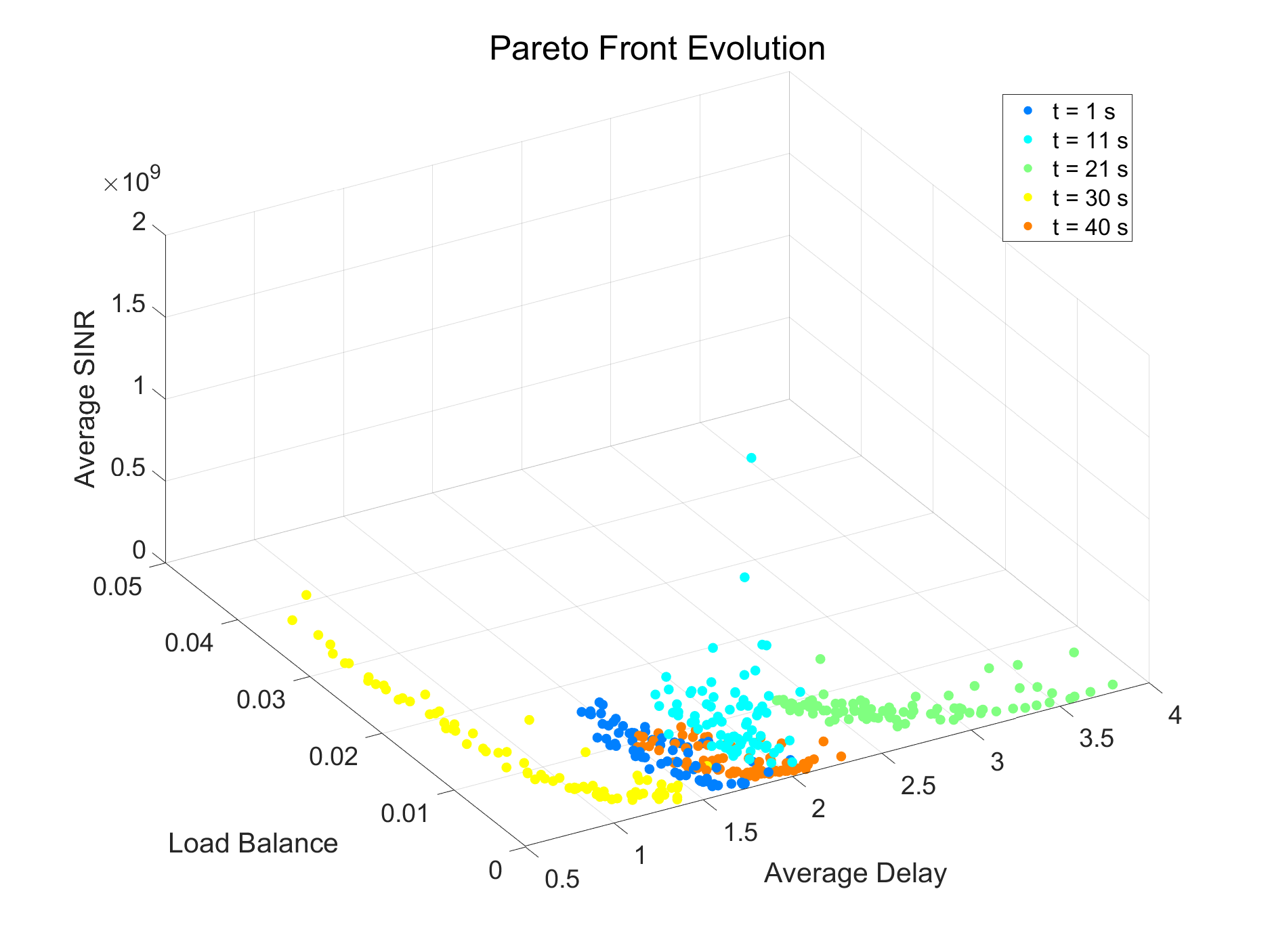}
			\caption{Inheritance Proportion: 80\%}
			\label{fig:pareto08_s2}
		\end{subfigure}
		\caption{Pareto Fronts with Different Inheritance Proportion on Scenario 2}
		\label{fig:pareto_fronts_s2}
	\end{figure*}

	The performance of the proposed algorithm in Scenario 2 is illustrated in Fig.~\ref{fig:pareto_fronts_s2}, where the Pareto fronts are presented across three objectives: Average Delay, Load Balance, and Average SINR. In this scenario, the vehicular density fluctuates between approximately 15 and 25 vehicles over the 40-second period, creating a dynamically changing optimization environment.	
	
	The Pareto fronts exhibit distinct characteristics under different inheritance proportions. Without inheritance (Fig.~\ref{fig:pareto00_s2}), the algorithm generates the most diverse solution set, forming a smooth and well-distributed Pareto surface. This indicates the algorithm's strong exploration capability in the absence of historical solutions. The solutions span broadly across all three objectives, with Average Delay ranging from 0.5 to 4, Load Balance from 0.005 to 0.02, and Average SINR reaching up to $15\times 10^8$.
	
	With a 30\% inheritance proportion (Fig.~\ref{fig:pareto03_s2}), the Pareto front maintains good diversity while showing initial signs of solution convergence. The surface remains continuous but exhibits slightly lower solution density compared to the no-inheritance case. This suggests that moderate inheritance helps focus the search while preserving sufficient exploration capability.
	
	As the inheritance proportion increases to 50\% (Fig.~\ref{fig:pareto05_s2}), the Pareto front becomes notably more concentrated, forming a distinctive band-like structure. The solutions show stronger convergence towards specific regions of the objective space, particularly in terms of Load Balance (0.008-0.015) and Average Delay (1.5-2.5). This indicates that higher inheritance begins to restrict the exploration of the solution space.
	
	At 80\% inheritance (Fig.~\ref{fig:pareto08_s2}), the Pareto front exhibits the highest degree of concentration, with solutions clustering tightly in a limited region of the objective space. While this demonstrates strong convergence properties, it also reveals that excessive inheritance may overly constrain the algorithm's ability to explore diverse trade-offs, particularly in dynamic scenarios where flexibility is crucial.
	
	The temporal evolution of solutions, indicated by different colors in the figures, reveals that higher inheritance proportions lead to more temporally consistent solution distributions, albeit at the cost of reduced diversity. This trade-off between solution diversity and temporal consistency becomes increasingly pronounced as the inheritance proportion increases.
	
	\begin{figure*}[!htbp]
		\centering
		% 第一个子图
		\begin{subfigure}{0.24\textwidth}
			\centering
			\includegraphics[width=\textwidth]{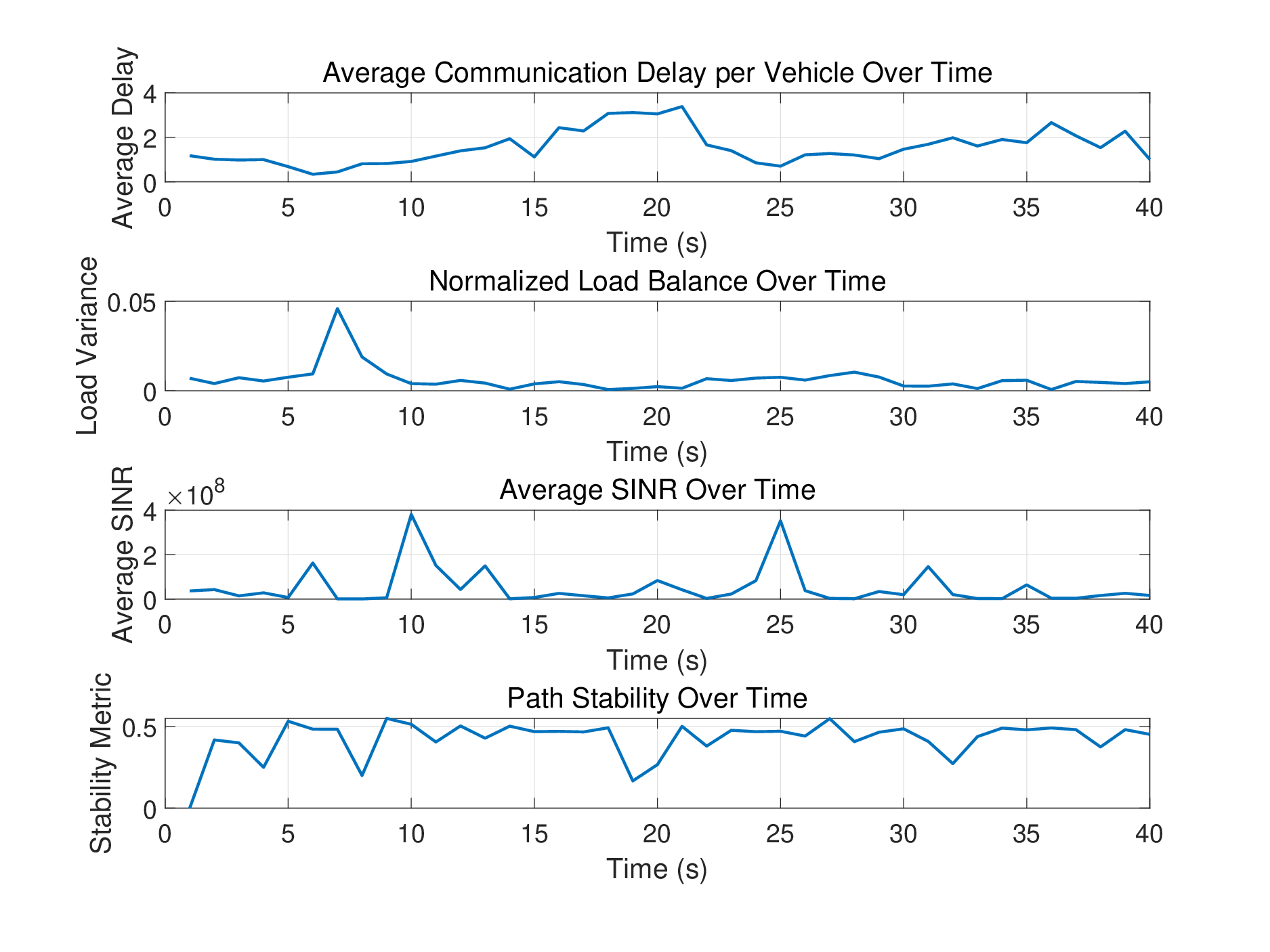}
			\caption{Performance with inheritance ratio: 0.0}
			\label{fig:perform00_s2}
		\end{subfigure}
		% 第二个子图
		\begin{subfigure}{0.24\textwidth}
			\centering
			\includegraphics[width=\textwidth]{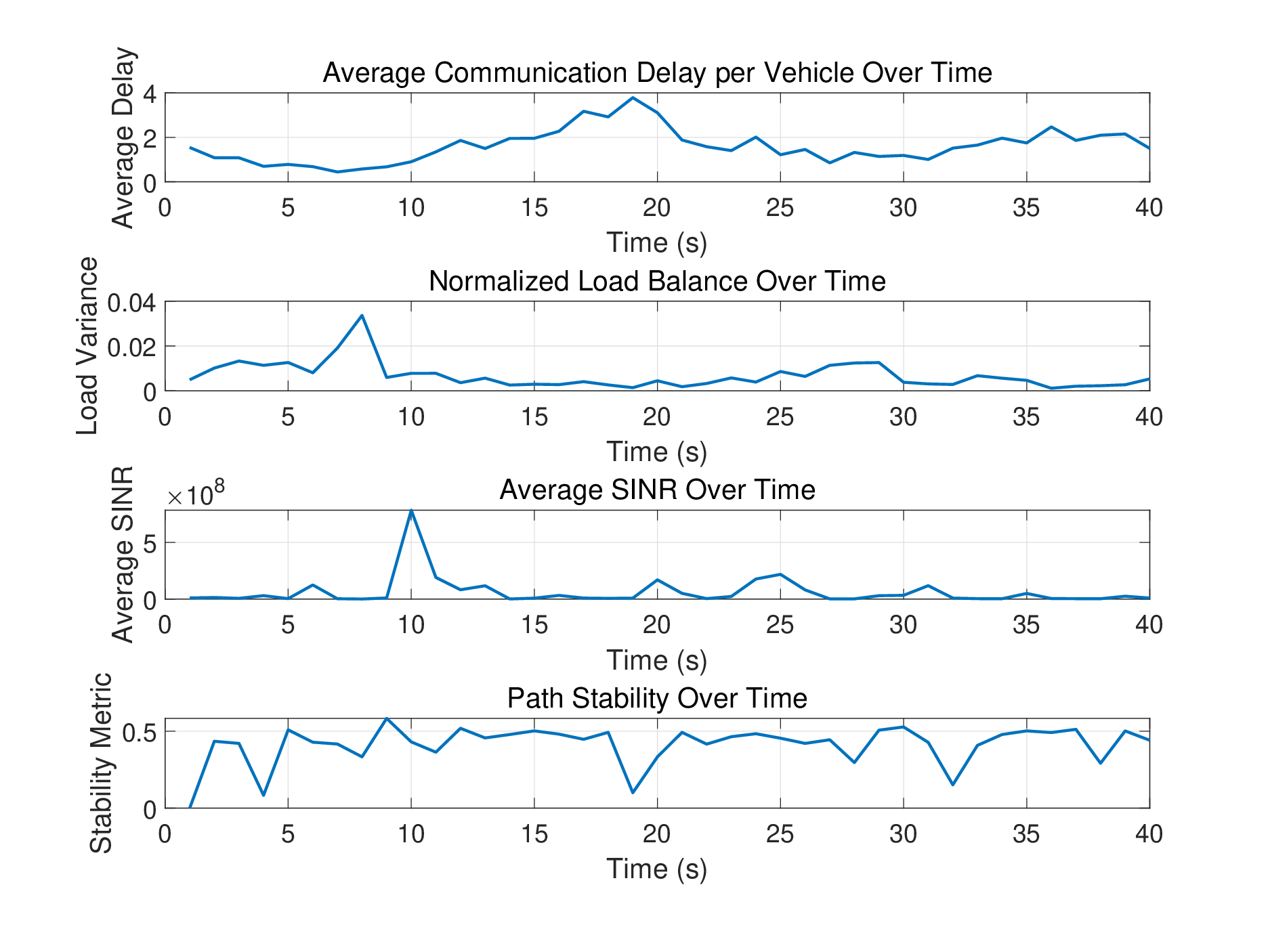}
			\caption{Performance with inheritance ratio: 0.3}
			\label{fig:perform03_s2}
		\end{subfigure}
		% 第三个子图
		\begin{subfigure}{0.24\textwidth}
			\centering
			\includegraphics[width=\textwidth]{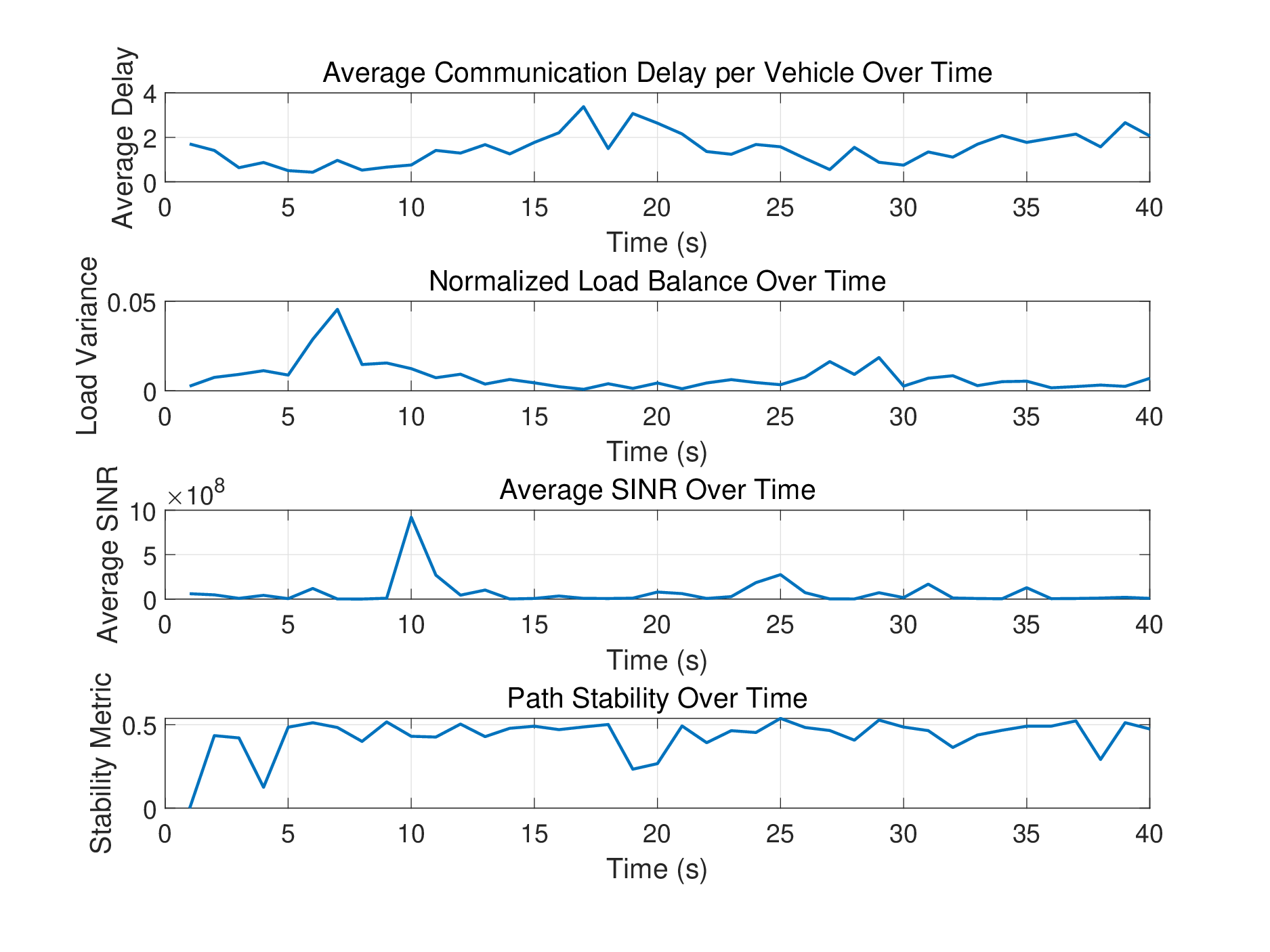}
			\caption{Performance with inheritance ratio: 0.5}
			\label{fig:perform05_s2}
		\end{subfigure}
		% 第三个子图
		\begin{subfigure}{0.24\textwidth}
			\centering
			\includegraphics[width=\textwidth]{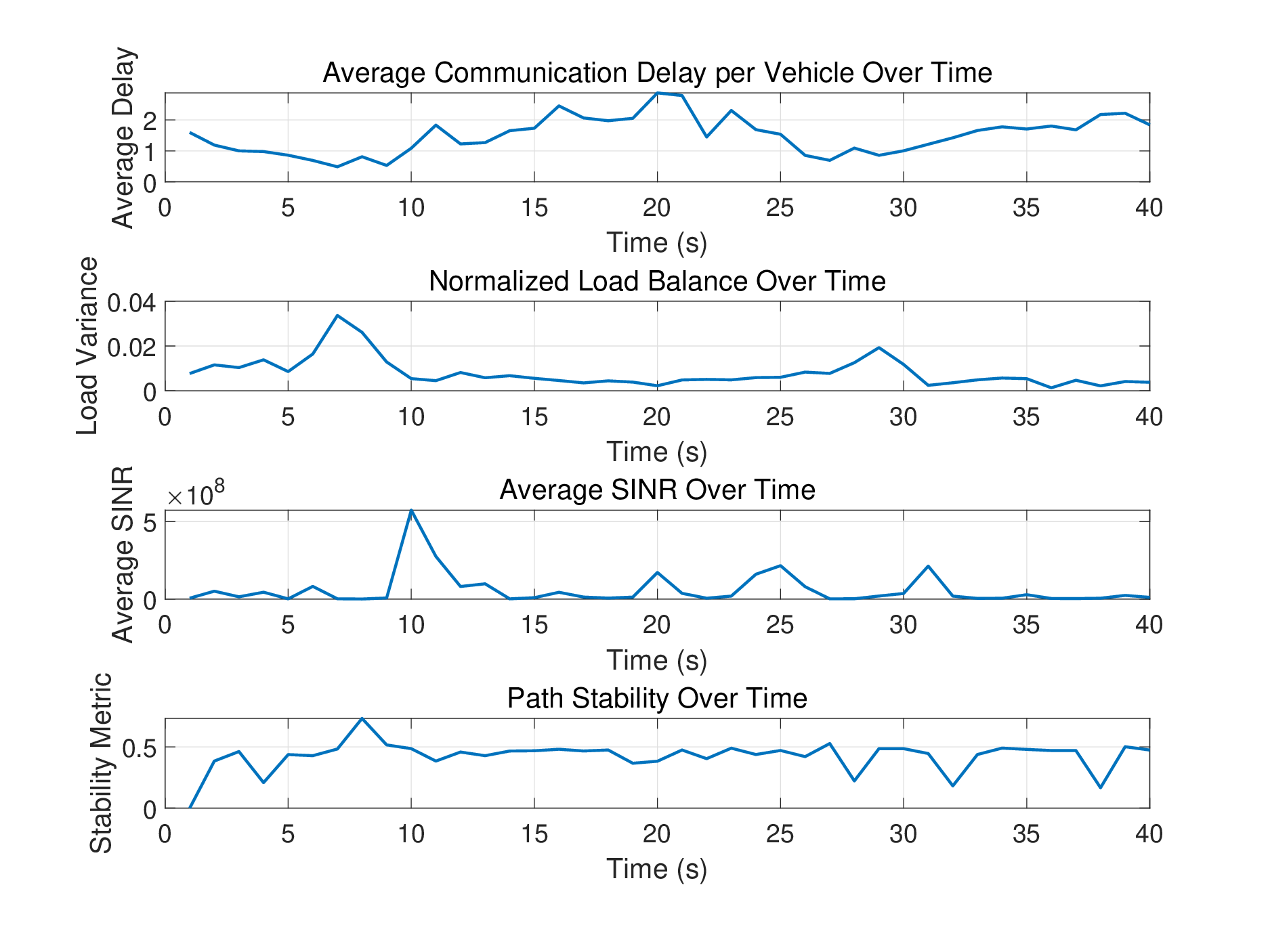}
			\caption{Performance with inheritance ratio: 0.8}
			\label{fig:perform08_s2}
		\end{subfigure}
		\caption{Performances of Different Inheritance Proportion on Scenario 2}
		\label{fig:performance_s2}
	\end{figure*}
	
	The performance metrics for different inheritance ratios are presented in Fig.~\ref{fig:performance_s2}, revealing distinctive patterns in system behavior across various metrics.
	
	Without inheritance (Fig.~\ref{fig:perform00_s2}), the Average Communication Delay maintains relatively low values around 2 units. The Normalized Load Balance exhibits fluctuations within a range of 0 to 0.05, while the Average SINR shows periodic spikes reaching up to $10\times 10^8$. Path Stability oscillates between 0 and 0.5, reflecting the system's continuous adaptation without historical influence.
	
	At an inheritance ratio of 0.3 (Fig.~\ref{fig:perform03_s2}), the Average Delay remains comparable to the zero-inheritance case, seldom exceeding 2.5 units. The Load Balance demonstrates similar variation patterns but with slightly reduced peak values. The SINR maintains periodic fluctuations with decreased peak magnitudes, while Path Stability shows comparable oscillation patterns to the zero-inheritance scenario.
	
	With an inheritance ratio of 0.5 (Fig.~\ref{fig:perform05_s2}), the Average Delay begins to increase, occasionally reaching 3 units. The Load Balance range expands to 0-0.04, indicating reduced load distribution efficiency. SINR peaks become more pronounced, particularly during certain time periods, while Path Stability maintains consistent oscillations with slightly improved baseline values.
	
	At the highest inheritance ratio of 0.8 (Fig.~\ref{fig:perform08_s2}), the system exhibits distinct performance characteristics. The Average Communication Delay fluctuates between 2 and 3 units, showing moderate variations but maintaining relatively stable bounds. While this performance is slightly higher than lower inheritance ratios, the difference is not as dramatic as previously suggested. The Average SINR demonstrates notably larger and more frequent fluctuations compared to other inheritance scenarios, indicating increased sensitivity in signal quality. The Load Balance remains contained within a similar range as the 0.5 inheritance case, while Path Stability shows comparable oscillation patterns with slight improvements in baseline stability.	
	
	This detailed analysis suggests that increasing inheritance ratios beyond 0.3 leads to deteriorating communication delay performance without proportional improvements in other metrics. The 0.3 inheritance ratio appears to offer the best balance, maintaining low delays while achieving comparable stability and load balance performance to higher inheritance ratios. These findings highlight the importance of carefully tuning inheritance proportions to optimize system performance in dynamic vehicular networks.
	
	\subsubsection{Scenario 3}
\begin{figure*}[!htbp]
		\centering
		% 第一个子图
		\begin{subfigure}{0.24\textwidth}
			\centering
			\includegraphics[width=\textwidth]{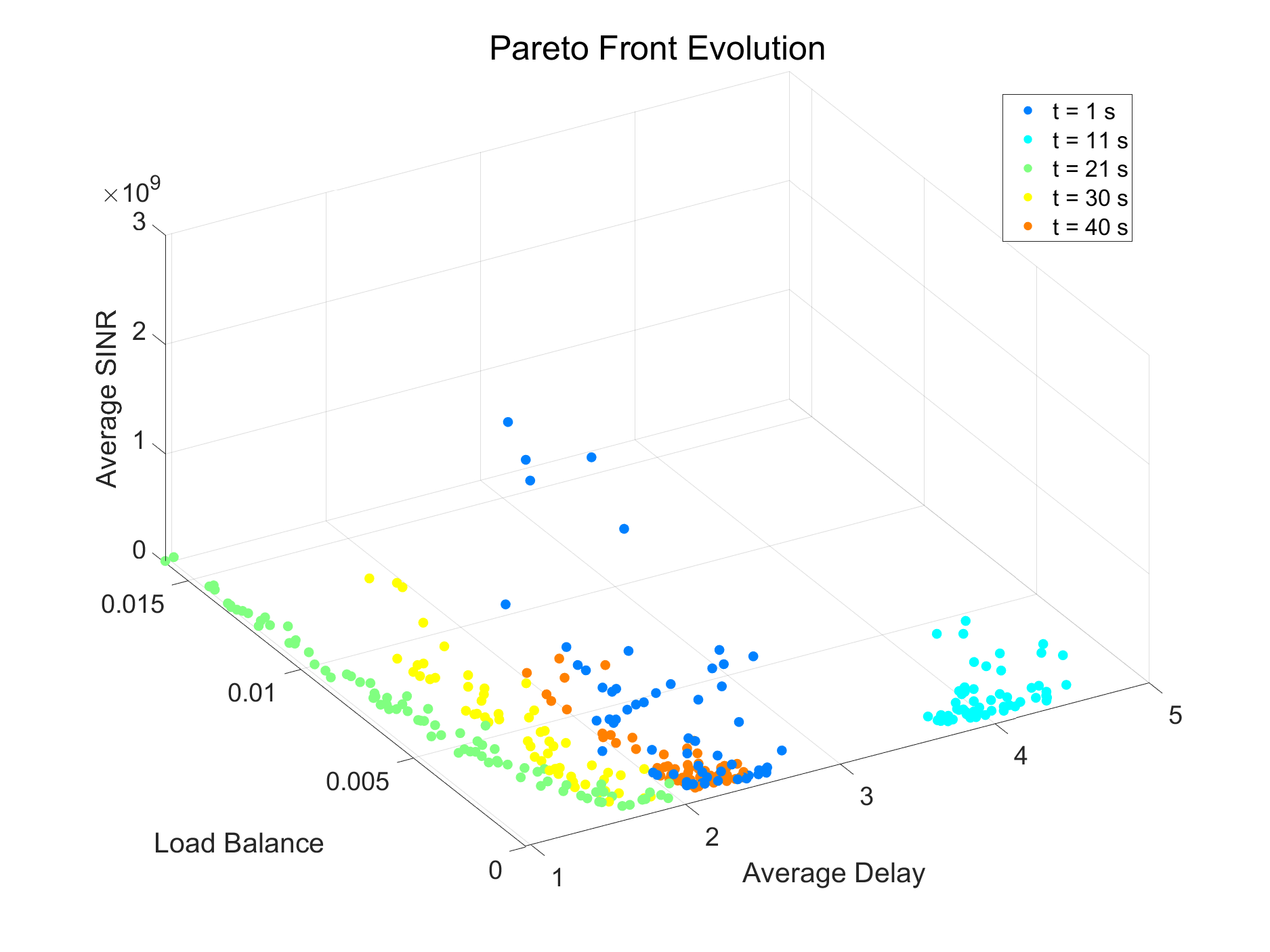}
			\caption{Inheritance Proportion: 0}
			\label{fig:pareto00_s3}
		\end{subfigure}
		% 第二个子图
		\begin{subfigure}{0.24\textwidth}
			\centering
			\includegraphics[width=\textwidth]{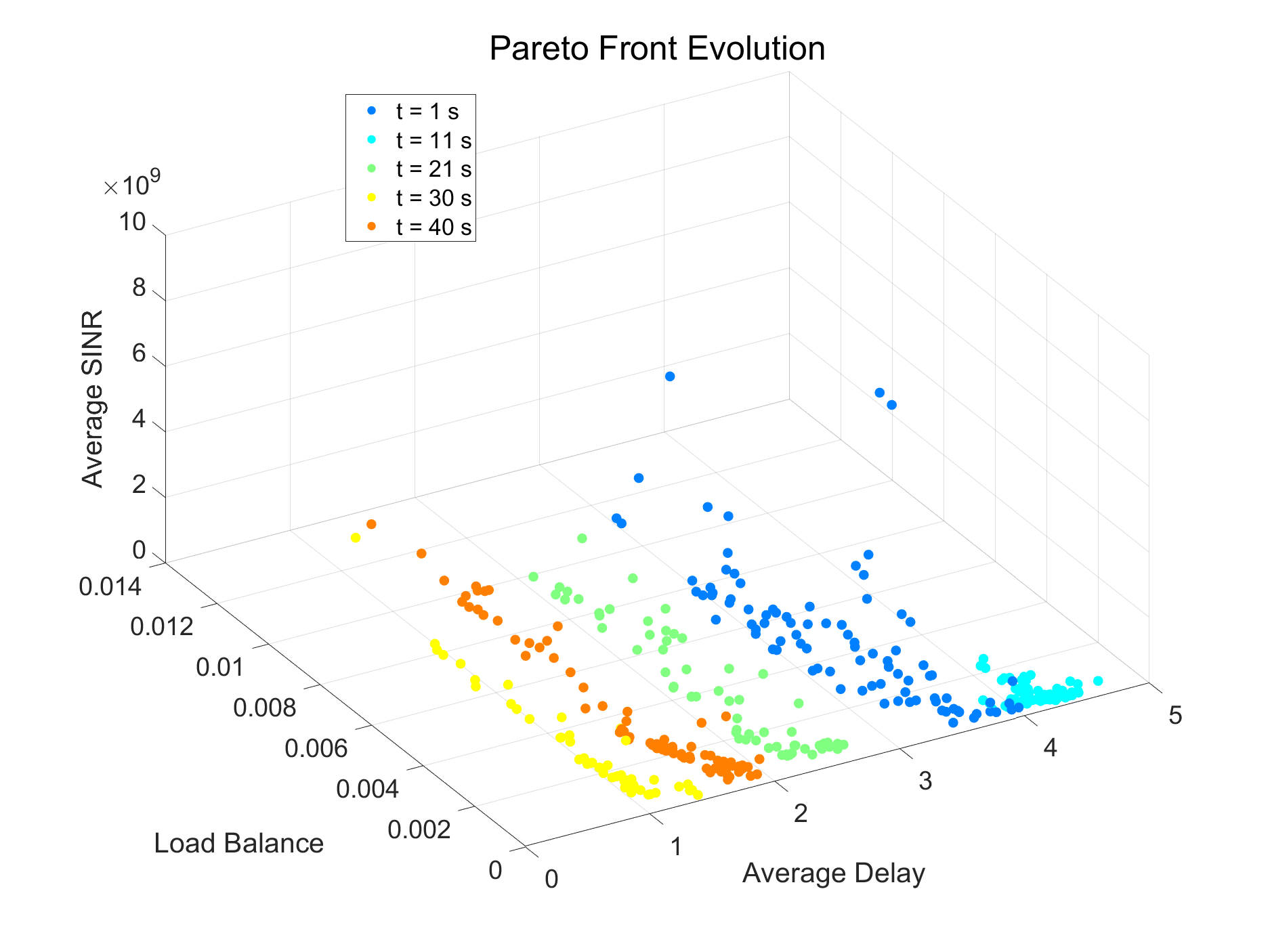}
			\caption{Inheritance Proportion: 30\%}
			\label{fig:pareto03_s3}
		\end{subfigure}
		% 第三个子图
		\begin{subfigure}{0.24\textwidth}
			\centering
			\includegraphics[width=\textwidth]{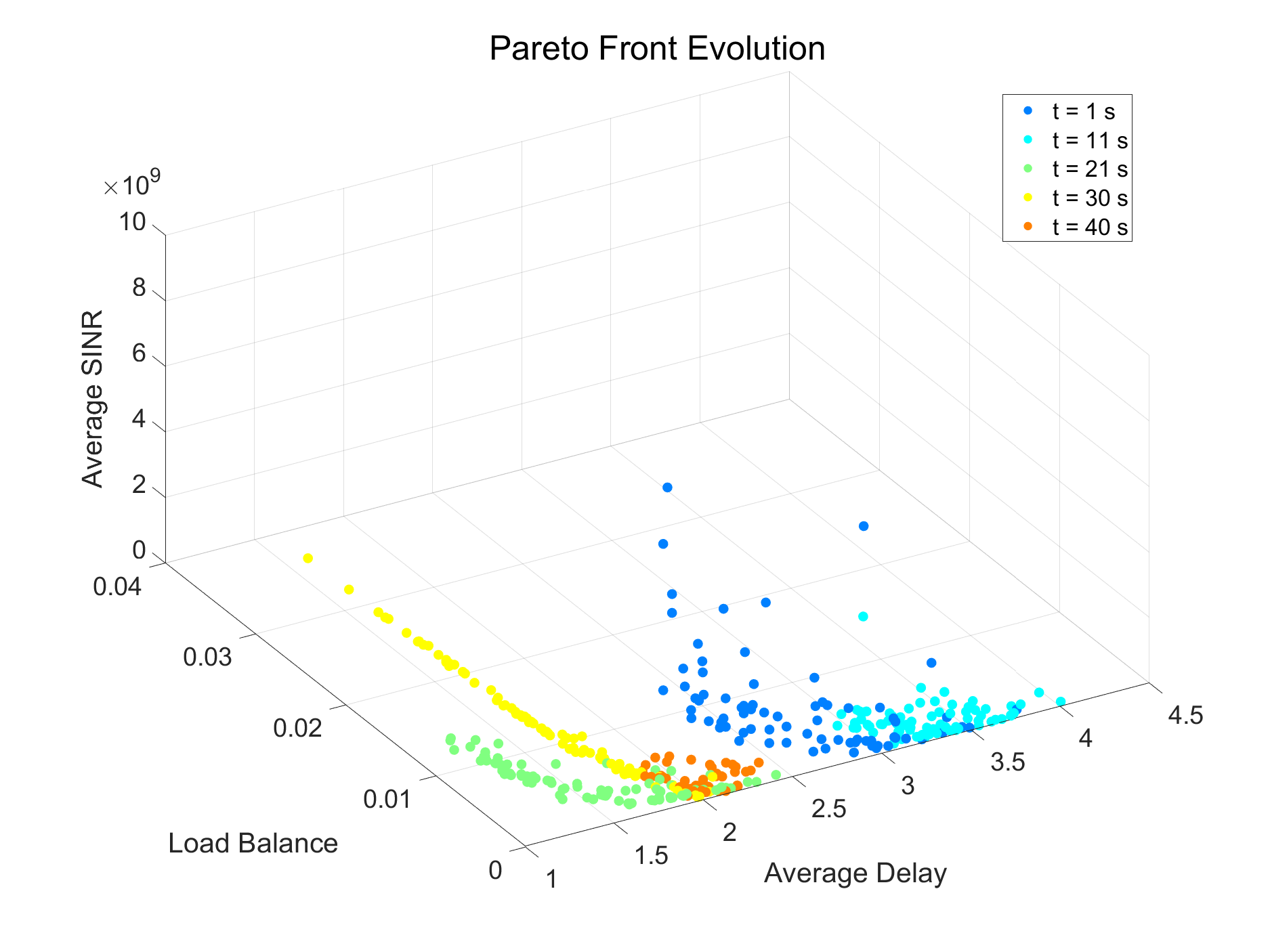}
			\caption{Inheritance Proportion: 50\%}
			\label{fig:pareto05_s3}
		\end{subfigure}
		% 第四个子图
		\begin{subfigure}{0.24\textwidth}
			\centering
			\includegraphics[width=\textwidth]{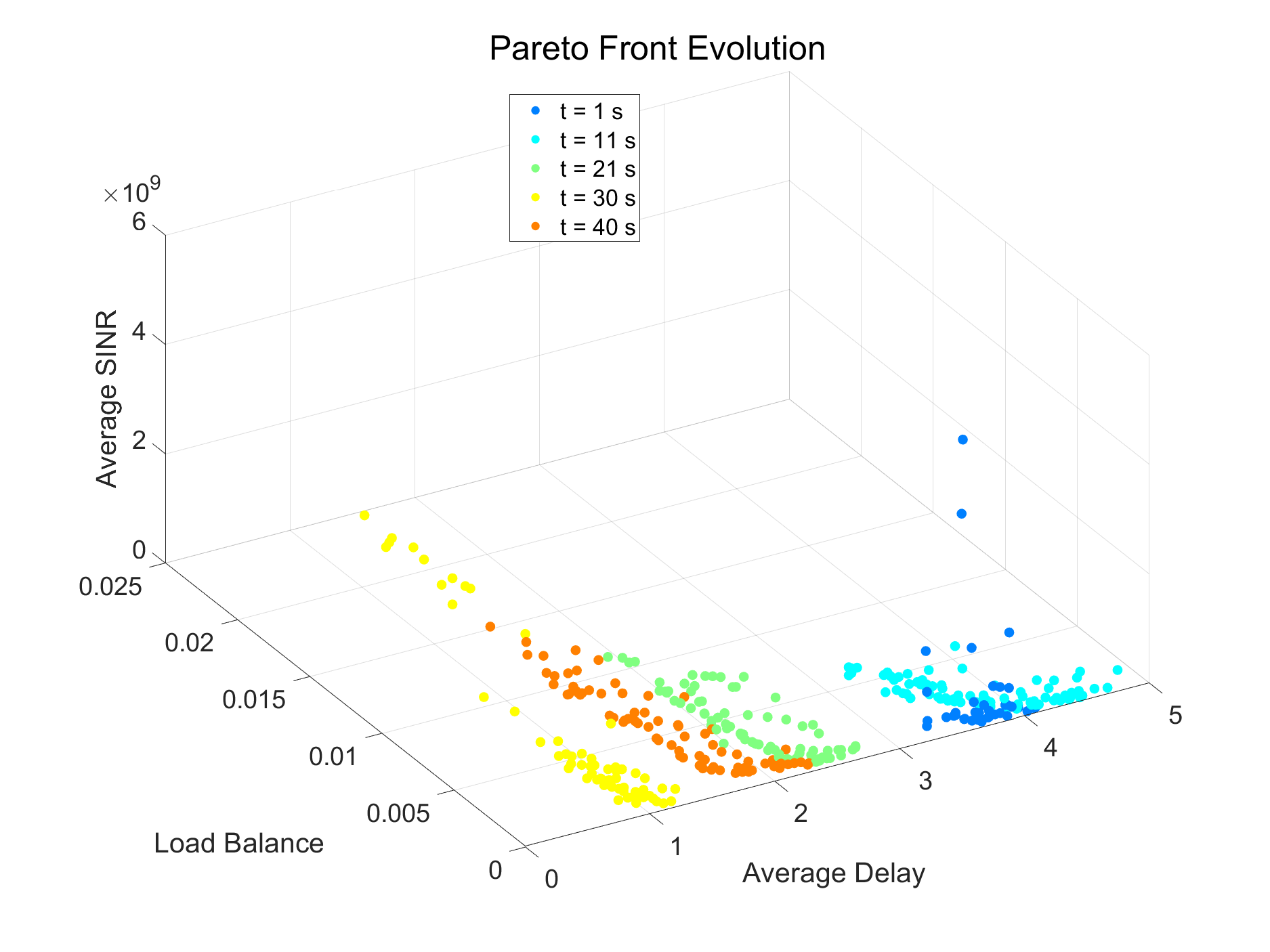}
			\caption{Inheritance Proportion: 80\%}
			\label{fig:pareto08_s3}
		\end{subfigure}
		\caption{Pareto Fronts with Different Inheritance Proportion on Scenario 3}
		\label{fig:pareto_fronts_s3}
	\end{figure*}

	Fig.~\ref{fig:pareto_fronts_s3} presents the evolution of Pareto fronts under different inheritance proportions, revealing distinct characteristics in solution distribution and objective trade-offs. The solutions are evaluated across three dimensions: Average Delay (0-5 units), Load Balance (0-0.015), and Average SINR (varying ranges up to $10\times 10^8$).

	Without inheritance (Fig.~\ref{fig:pareto00_s3}), the algorithm generates a relatively complete Pareto surface, though with noticeable clustering in certain regions. The solutions span across the entire Average Delay range but show limited diversity in the SINR dimension, reaching only up to $3\times 10^8$. The temporal evolution of solutions, indicated by different colors, reveals certain discontinuities in the front's development.
	
	With 30\% inheritance (Fig.~\ref{fig:pareto03_s3}), the Pareto front exhibits the most uniform distribution among all cases, with solutions well-spread across the objective space. Notably, the SINR range expands significantly, reaching up to $10\times 10^8$, while maintaining good coverage in both Average Delay and Load Balance dimensions. The temporal progression of solutions shows consistent exploration across different time periods, indicating robust adaptation capability.
	
	At 50\% inheritance (Fig.~\ref{fig:pareto05_s3}), the solutions begin to show moderate convergence behavior while maintaining reasonable diversity. The Pareto front remains continuous but demonstrates a preference for intermediate delay values (2.5-4 units) and moderate Load Balance levels. The SINR range remains substantial but shows more concentrated distribution compared to the 30\% case.
	
	The 80\% inheritance ratio (Fig.~\ref{fig:pareto08_s3}) leads to more pronounced solution clustering, particularly in the Average Delay range of 2-3 units. While this indicates strong convergence properties, it also suggests reduced exploration capability. The SINR range contracts notably, rarely exceeding $6\times10^8$, and solutions exhibit tighter grouping in the Load Balance dimension.
	
	This comparative analysis reveals that 30\% inheritance achieves the most favorable balance between solution diversity and convergence, facilitating comprehensive exploration of the objective space while maintaining solution quality. Higher inheritance ratios progressively constrain the search space, potentially limiting the algorithm's ability to discover diverse trade-offs in dynamic network environments.
	
	\begin{figure*}[!htbp]
		\centering
		% 第一个子图
		\begin{subfigure}{0.24\textwidth}
			\centering
			\includegraphics[width=\textwidth]{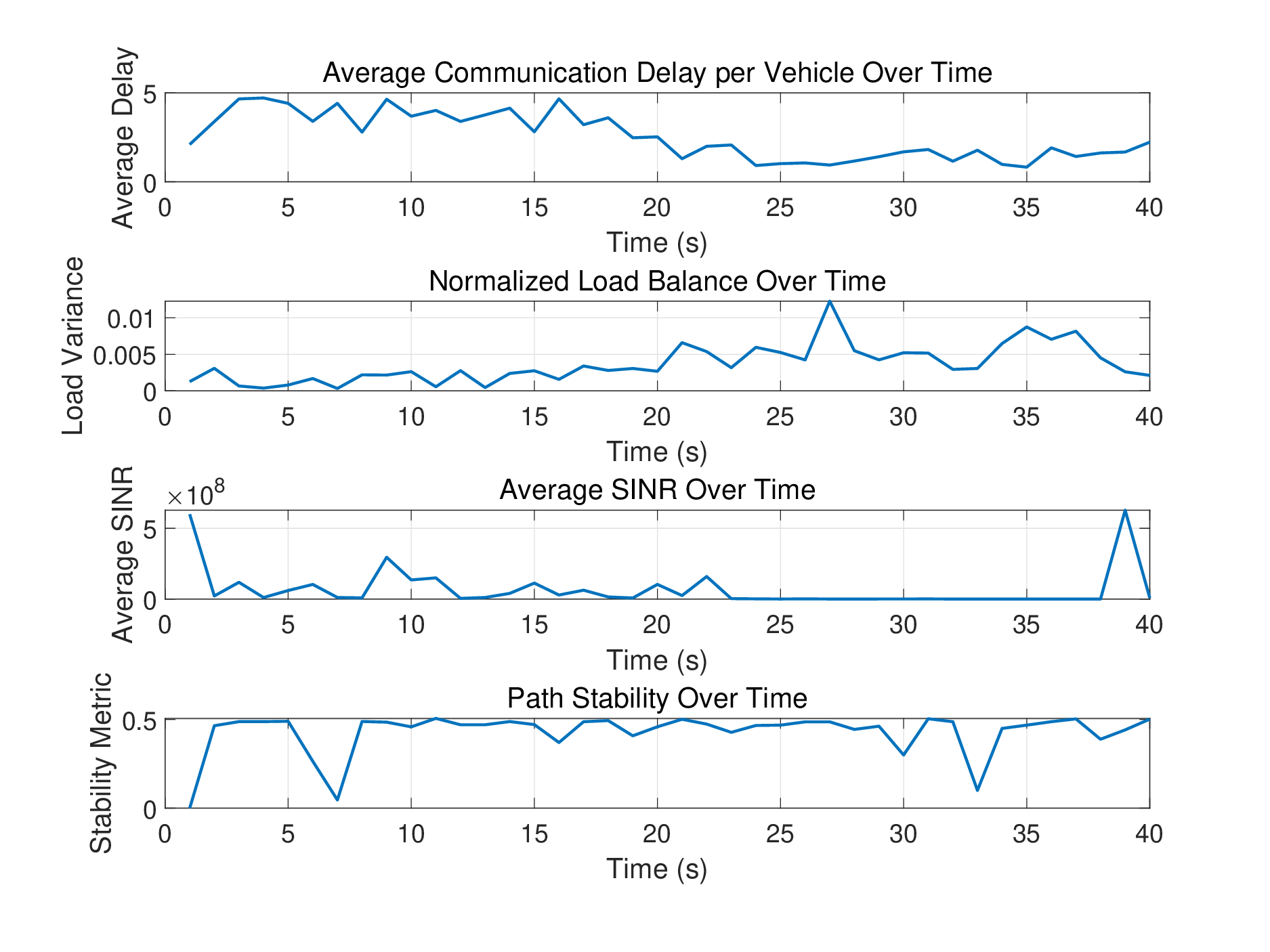}
			\caption{Performance with inheritance ratio: 0.0}
			\label{fig:perform00_s3}
		\end{subfigure}
		% 第二个子图
		\begin{subfigure}{0.24\textwidth}
			\centering
			\includegraphics[width=\textwidth]{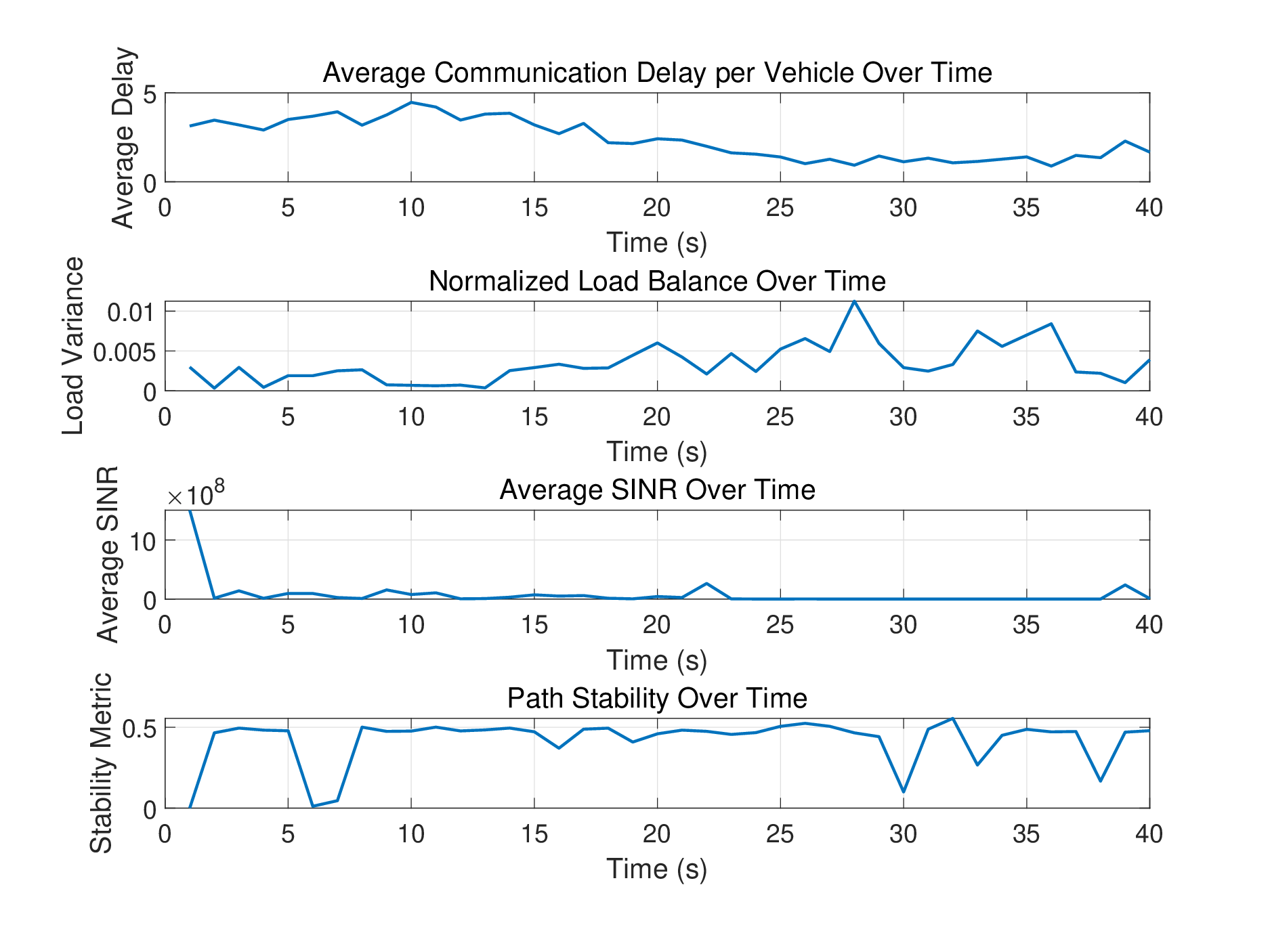}
			\caption{Performance with inheritance ratio: 0.3}
			\label{fig:perform03_s3}
		\end{subfigure}
		% 第三个子图
		\begin{subfigure}{0.24\textwidth}
			\centering
			\includegraphics[width=\textwidth]{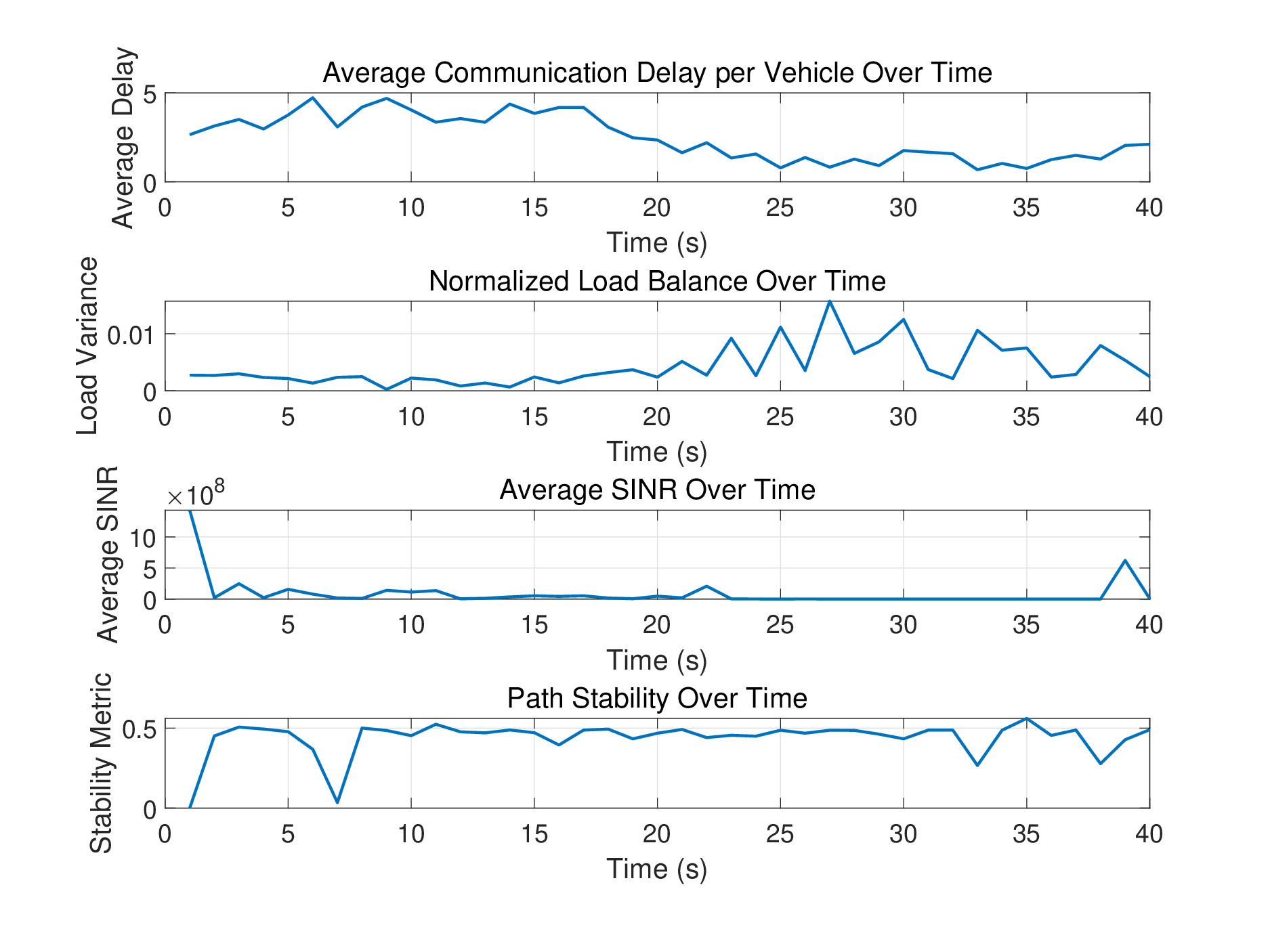}
			\caption{Performance with inheritance ratio: 0.5}
			\label{fig:perform05_s3}
		\end{subfigure}
		% 第四个子图
		\begin{subfigure}{0.24\textwidth}
			\centering
			\includegraphics[width=\textwidth]{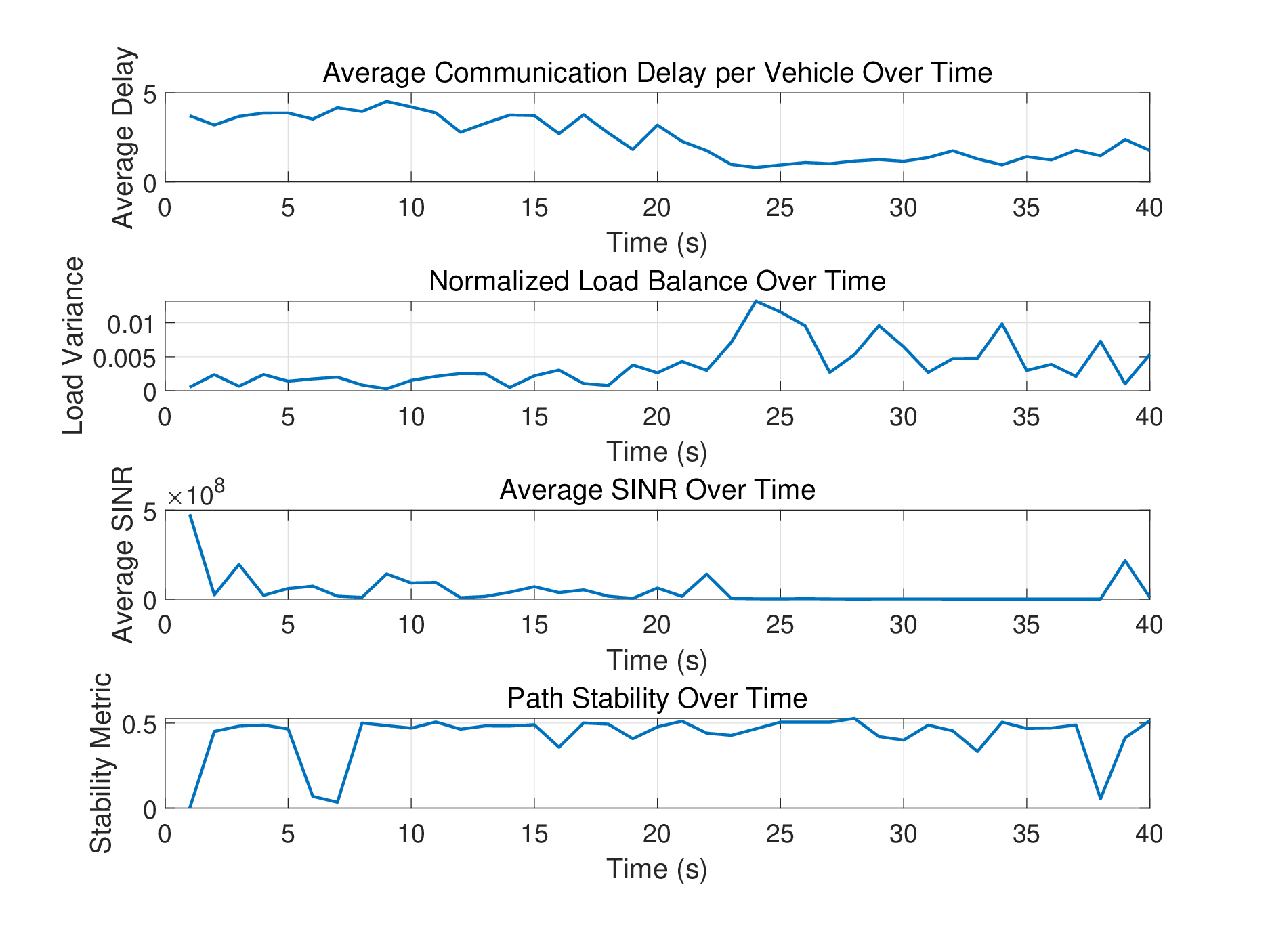}
			\caption{Performance with inheritance ratio: 0.8}
			\label{fig:perform08_s3}
		\end{subfigure}
		\caption{Performances of Different Inheritance Proportion on Scenario 3}
		\label{fig:performance_s3}
	\end{figure*}
	
	The performance metrics for different inheritance ratios in this scenario are presented in Fig.~\ref{fig:performance_s3}, demonstrating distinct patterns across various performance indicators.
	
	In the zero-inheritance case (Fig.~\ref{fig:perform00_s3}), the Average Communication Delay maintains values between 1.5 and 2.5 units with moderate fluctuations. The Normalized Load Balance exhibits small variations within 0-0.01, while the Average SINR shows periodic variations with a notable increase in the final period. Path Stability oscillates between 0 and 0.5, reflecting the algorithm's adaptation behavior without historical influence.
	
	With an inheritance ratio of 0.3 (Fig.~\ref{fig:perform03_s3}), the system demonstrates notably improved stability across all metrics. The Average Delay shows the smallest fluctuations among all cases, maintaining consistent values around 2 units. Load Balance variations are minimized, and SINR fluctuations are more controlled, particularly during the middle period. Path Stability exhibits a more regular pattern with fewer sharp transitions.
	
	At an inheritance ratio of 0.5 (Fig.~\ref{fig:perform05_s3}), the system maintains reasonable performance but with increased variations compared to the 0.3 case. The Average Delay shows slightly larger fluctuations, while Load Balance remains within similar bounds. SINR demonstrates more pronounced variations, particularly in the latter half of the time period, though the general pattern remains stable.
	
	The highest inheritance ratio of 0.8 (Fig.~\ref{fig:perform08_s3}) shows comparable Average Delay performance to other cases, contrary to expectations of significant deterioration. However, both SINR and Load Balance exhibit increased volatility, particularly in SINR measurements during the latter period. Path Stability maintains similar patterns to other inheritance ratios, suggesting that higher inheritance does not necessarily translate to improved stability in this scenario.
	
	The analysis reveals that a 0.3 inheritance ratio provides the most balanced performance, particularly in terms of delay stability and load balance. This observation differs from previous scenarios, highlighting the importance of context-specific optimization in dynamic vehicular networks. Higher inheritance ratios, while not detrimental to delay performance, show increased volatility in other metrics, suggesting that moderate inheritance better serves the system's overall stability and performance requirements.
	
	\subsection{Summary}
	In summary, we evaluated the proposed algorithm's performance across three distinct vehicular network scenarios, examining four inheritance ratios (0.0, 0.3, 0.5, and 0.8) through key performance metrics: average communication delay, normalized load balance, average SINR, and path stability.
	
	The results consistently demonstrate that a moderate inheritance ratio of 0.3 achieves optimal balance between solution diversity and convergence across all scenarios. While higher inheritance ratios enhance temporal consistency, they constrain the algorithm's exploration capabilities and increase performance volatility. The Pareto front analysis reveals that the 0.3 inheritance ratio effectively maintains solution diversity while providing sufficient historical information for optimization, making it particularly suitable for dynamic vehicular networks where both solution quality and adaptability are essential.	These findings underscore the importance of appropriate inheritance ratio selection in balancing computational efficiency and optimization performance in real-world vehicular network applications.

	\section{Conclusions and Future Work}
	\label{sec:con}
	This paper has addressed the critical challenge of maintaining reliable communication in dynamic vehicular networks through a novel multi-objective optimization framework that incorporates temporal continuity. The proposed framework successfully balances multiple competing objectives - communication delay, load distribution, link quality, and temporal stability - while adapting to rapidly changing network topologies. Through the integration of temporal awareness into the optimization process, our approach provides a more practical solution for real-world vehicular networks.
	
	The enhanced NSGA-II algorithm developed in this work demonstrates effective performance in handling dynamic optimization scenarios through several key innovations. The incorporation of dimension adaptation mechanisms allows seamless handling of varying vehicle counts, while the temporal inheritance strategy maintains solution quality across consecutive time steps. The normalized crowding distance calculation ensures balanced consideration of objectives operating at different scales, contributing to diverse and well-distributed Pareto-optimal solutions.
	
	Extensive experimentation across three distinct traffic scenarios has validated the effectiveness of our approach. The results consistently demonstrate that a moderate inheritance ratio of 0.3 achieves optimal balance between solution diversity and convergence. This inheritance ratio maintains good performance in communication delay and load balance while providing sufficient exploration capability for adapting to network changes. Higher inheritance ratios, while enhancing temporal consistency, tend to constrain the algorithm's exploration capabilities and increase performance volatility, particularly in SINR measurements.
	
	The Pareto front analysis reveals a fundamental trade-off between solution diversity and temporal consistency as the inheritance ratio increases. Without inheritance, the algorithm generates diverse but potentially disconnected solution sets, while higher inheritance ratios produce more concentrated but potentially limited solution spaces. The framework exhibits robust adaptation to both gradual and fluctuating changes in vehicle density, as demonstrated across different scenarios.
	
	Future research directions could explore adaptive inheritance ratio mechanisms that dynamically adjust based on the rate of topology change and optimization objectives. Additionally, investigating the framework's applicability to heterogeneous vehicle networks, incorporating different communication capabilities and requirements, would extend its practical utility. The integration of machine learning techniques to predict network topology changes could further enhance the optimization process. Real-world implementation studies focusing on computational efficiency and scalability would provide valuable insights for practical deployment in intelligent transportation systems.

	\bibliographystyle{IEEEtran}  % 或其他样式，如 plain、unsrt 等
	\bibliography{refe.bib}

\begin{thebibliography}{10}
\providecommand{\url}[1]{#1}
\csname url@samestyle\endcsname
\providecommand{\newblock}{\relax}
\providecommand{\bibinfo}[2]{#2}
\providecommand{\BIBentrySTDinterwordspacing}{\spaceskip=0pt\relax}
\providecommand{\BIBentryALTinterwordstretchfactor}{4}
\providecommand{\BIBentryALTinterwordspacing}{\spaceskip=\fontdimen2\font plus
\BIBentryALTinterwordstretchfactor\fontdimen3\font minus
  \fontdimen4\font\relax}
\providecommand{\BIBforeignlanguage}[2]{{%
\expandafter\ifx\csname l@#1\endcsname\relax
\typeout{** WARNING: IEEEtran.bst: No hyphenation pattern has been}%
\typeout{** loaded for the language `#1'. Using the pattern for}%
\typeout{** the default language instead.}%
\else
\language=\csname l@#1\endcsname
\fi
#2}}
\providecommand{\BIBdecl}{\relax}
\BIBdecl

\bibitem{maram_bani_younes_59aa1648}
M.~B. Younes and A.~Boukerche, ``Traffic efficiency applications over downtown
  roads: A new challenge for intelligent connected vehicles,'' \emph{ACM
  Computing Surveys}, vol.~53, no.~5, Sep. 2020.

\bibitem{eshita_rastogi_b566e6c6}
E.~Rastogi, M.~K. Maheshwari, A.~Roy, N.~Saxena, and D.~R. Shin, ``A novel
  safety message dissemination framework in lte-v2x system,''
  \emph{Transactions on Emerging Telecommunications Technologies}, vol.~32,
  no.~9, p. e4275, 2021.

\bibitem{craig_cooper_aa6ee789}
C.~Cooper, D.~Franklin, M.~Ros, F.~Safaei, and M.~Abolhasan, ``A comparative
  survey of vanet clustering techniques,'' \emph{IEEE Communications Surveys \&
  Tutorials}, vol.~19, no.~1, pp. 657--681, 2017.

\bibitem{junjie_zhang_fe5a8711}
J.~Zhang, M.~Ye, Z.~Guo, C.-Y. Yen, and H.~J. Chao, ``Cfr-rl: Traffic
  engineering with reinforcement learning in sdn,'' \emph{IEEE Journal on
  Selected Areas in Communications}, vol.~38, no.~10, pp. 2249--2259, 2020.

\bibitem{md__noor_a_rahim_4b96d3b0}
M.~Noor-A-Rahim, Z.~Liu, H.~Lee, G.~G. M.~N. Ali, D.~Pesch, and P.~Xiao, ``A
  survey on resource allocation in vehicular networks,'' \emph{IEEE
  Transactions on Intelligent Transportation Systems}, vol.~23, no.~2, pp.
  701--721, 2022.

\bibitem{xianghui_cao_b8d3413d}
X.~Cao, L.~Liu, Y.~Cheng, L.~X. Cai, and C.~Sun, ``On optimal device-to-device
  resource allocation for minimizing end-to-end delay in vanets,'' \emph{IEEE
  Transactions on Vehicular Technology}, vol.~65, no.~10, pp. 7905--7916, 2016.

\bibitem{wantanee_viriyasitavat_f5df62bb}
W.~Viriyasitavat, M.~Boban, H.-M. Tsai, and A.~Vasilakos, ``Vehicular
  communications: Survey and challenges of channel and propagation models,''
  \emph{IEEE Vehicular Technology Magazine}, vol.~10, no.~2, pp. 55--66, 2015.

\bibitem{kai_xiong_ee6b167b}
K.~Xiong, S.~Leng, X.~Chen, C.~Huang, C.~Yuen, and Y.~L. Guan, ``Communication
  and computing resource optimization for connected autonomous driving,''
  \emph{IEEE Transactions on Vehicular Technology}, vol.~69, no.~11, pp.
  12\,652--12\,663, 2020.

\bibitem{huacheng_zeng_7b63c042}
H.~Zeng, H.~Pirayesh, P.~K. Sangdeh, and A.~Quadri, ``Vehcom: Delay-guaranteed
  message broadcast for large-scale vehicular networks,'' \emph{IEEE
  Transactions on Wireless Communications}, vol.~20, no.~6, pp. 3883--3896,
  2021.

\bibitem{francesco_malandrino_46ebdf5f}
F.~Malandrino, C.~F. Chiasserini, and G.~M. Dell’Aera, ``Edge-powered
  assisted driving for connected cars,'' \emph{IEEE Transactions on Mobile
  Computing}, vol.~22, no.~2, pp. 874--889, 2023.

\bibitem{jiawen_kang_945bf2ac}
J.~Kang, J.~He, H.~Du, Z.~Xiong, Z.~Yang, X.~Huang, and S.~Xie, ``Adversarial
  attacks and defenses for semantic communication in vehicular metaverses,''
  \emph{IEEE Wireless Communications}, vol.~30, no.~4, pp. 48--55, 2023.

\bibitem{zhonghui_pei_90e7494f}
Z.~Pei, W.~Chen, H.~Zheng, and L.~Du, ``Optimization of maximum routing hop
  count parameter based on vehicle density for vanet,'' \emph{Mobile
  Information Systems}, vol. 2020, no.~1, p. 2741648, 2020.

\bibitem{jin_tian_83309c8b}
J.~Tian and F.~Meng, ``Comparison survey of mobility models in vehicular ad-hoc
  network (vanet),'' in \emph{2020 IEEE 3rd International Conference on
  Automation, Electronics and Electrical Engineering (AUTEEE)}, 2020, pp.
  337--342.

\bibitem{jeng_ji_huang_855b9d66}
J.-J. Huang and Y.-T. Tseng, ``The steady-state distribution of rehealing delay
  in an intermittently connected highway vanet,'' \emph{IEEE Transactions on
  Vehicular Technology}, vol.~67, no.~10, pp. 10\,010--10\,021, 2018.

\bibitem{felipe_d__cunha_945016f9}
F.~Cunha, L.~Villas, A.~Boukerche, G.~Maia, A.~Viana, R.~A. Mini, and A.~A.
  Loureiro, ``Data communication in vanets: Protocols, applications and
  challenges,'' \emph{Ad Hoc Networks}, vol.~44, pp. 90--103, 2016.

\bibitem{silviu_andrei_lazar_493c4ad6}
S.-A. Lazar, M.~K. Hamadani, and C.-E. Stefan, ``Optimization analysis of
  vanet's control plane for safety application traffic,'' in \emph{2018
  International Conference on Communications (COMM)}, 2018, pp. 305--308.

\bibitem{ralf_schmitz_f52c4b83}
R.~Schmitz, A.~Leiggener, A.~Festag, L.~Eggert, and W.~Effelsberg, ``Analysis
  of path characteristics and transport protocol design in vehicular ad hoc
  networks,'' in \emph{2006 IEEE 63rd Vehicular Technology Conference}, vol.~2,
  2006, pp. 528--532.

\bibitem{amina_bengag_31f1a673}
B.~Amina and E.~Mohamed, ``Performance evaluation of vanets routing protocols
  using sumo and ns3,'' in \emph{2018 IEEE 5th International Congress on
  Information Science and Technology (CiSt)}, 2018, pp. 525--530.

\bibitem{antonio_russoniello_44f40609}
A.~Russoniello and E.~Gamess, ``Evaluation of different routing protocols for
  mobile ad-hoc networks in scenarios with high-speed mobility,''
  \emph{International Journal of Computer Network and Information Security
  (IJCNIS)}, vol.~10, no.~10, pp. 46--52, 2018.

\bibitem{jiayue_he_ed1c7260}
J.~He, M.~Bresler, M.~Chiang, and J.~Rexford, ``Towards robust multi-layer
  traffic engineering: Optimization of congestion control and routing,''
  \emph{IEEE Journal on Selected Areas in Communications}, vol.~25, no.~5, pp.
  868--880, 2007.

\bibitem{xiaoyun_xie_85881fa0}
X.~Xie, Y.~D. Navaei, and S.~Einy, ``A clustering-based routing protocol using
  path pattern discovery method to minimize delay in vanet,'' \emph{Wireless
  Communications and Mobile Computing}, vol. 2023, no.~1, p. 3776815, 2023.

\bibitem{lei_zhang_bfc944ca}
L.~Zhang, Y.~Cui, M.~Wang, K.~Zhu, Y.~Zhu, and Y.~Jiang, ``Deepcc: Bridging the
  gap between congestion control and applications via multiobjective
  optimization,'' \emph{IEEE/ACM Transactions on Networking}, vol.~30, no.~5,
  pp. 2274--2288, 2022.

\bibitem{yasar_sinan_nasir_9cb0d554}
Y.~S. Nasir and D.~Guo, ``Deep reinforcement learning for joint spectrum and
  power allocation in cellular networks,'' in \emph{2021 IEEE Globecom
  Workshops (GC Wkshps)}, 2021, pp. 1--6.

\bibitem{waleed_ejaz_49205af4}
W.~Ejaz, S.~K. Sharma, S.~Saadat, M.~Naeem, A.~Anpalagan, and N.~Chughtai, ``A
  comprehensive survey on resource allocation for cran in 5g and beyond
  networks,'' \emph{Journal of Network and Computer Applications}, vol. 160, p.
  102638, 2020.

\bibitem{rui_dong_e7e13766}
R.~Dong, C.~She, W.~Hardjawana, Y.~Li, and B.~Vucetic, ``Deep learning for
  radio resource allocation with diverse quality-of-service requirements in
  5g,'' \emph{IEEE Transactions on Wireless Communications}, vol.~20, no.~4,
  pp. 2309--2324, 2021.

\bibitem{fei_song_badc5f02}
F.~Song, J.~Li, C.~Ma, Y.~Zhang, L.~Shi, and D.~N.~K. Jayakody, ``Dynamic
  virtual resource allocation for 5g and beyond network slicing,'' \emph{IEEE
  Open Journal of Vehicular Technology}, vol.~1, pp. 215--226, 2020.

\bibitem{linlin_sun_b468e9a7}
L.~Sun, T.~Xu, S.~Yan, J.~Hu, X.~Yu, and F.~Shu, ``On resource allocation in
  covert wireless communication with channel estimation,'' \emph{IEEE
  Transactions on Communications}, vol.~68, no.~10, pp. 6456--6469, 2020.

\bibitem{cao_chen_60c87af1}
C.~Chen, F.~Zhou, M.~Tornatore, and S.~Xiao, ``Maximizing revenue with adaptive
  modulation and multiple fecs in flexible optical networks,'' \emph{IEEE/ACM
  Transactions on Network}, vol.~31, no.~1, p. 220–233, Aug. 2022.

\bibitem{anirudh_subramanyam_78638638}
A.~Subramanyam, P.~P. Repoussis, and C.~E. Gounaris, ``Robust optimization of a
  broad class of heterogeneous vehicle routing problems under demand
  uncertainty,'' \emph{Institute for Operations Research and the Management
  Sciences}, vol.~32, no.~3, pp. 661--681, 01 2020.

\bibitem{gagan_preet_kour_marwah_d7232a31}
G.~P.~K. Marwah and A.~Jain, ``A hybrid optimization with ensemble learning to
  ensure vanet network stability based on performance analysis,''
  \emph{Scientific Reports}, vol.~12, no.~1, p. 10287, 2022.

\bibitem{weihua_wu_46918094}
W.~Wu, R.~Liu, Q.~Yang, and T.~Q.~S. Quek, ``Robust resource allocation for
  vehicular communications with imperfect csi,'' \emph{IEEE Transactions on
  Wireless Communications}, vol.~20, no.~9, pp. 5883--5897, 2021.

\bibitem{yiannos_mylonas_00b9ff6f}
Y.~Mylonas, M.~Lestas, A.~Pitsillides, P.~Ioannou, and V.~Papadopoulou, ``Speed
  adaptive probabilistic flooding for vehicular ad hoc networks,'' \emph{IEEE
  Transactions on Vehicular Technology}, vol.~64, no.~5, pp. 1973--1990, 2015.

\bibitem{jingqiu_guo_7aa54d64}
J.~Guo, Y.~Zhang, X.~Chen, S.~Yousefi, C.~Guo, and Y.~Wang, ``Spatial
  stochastic vehicle traffic modeling for vanets,'' \emph{IEEE Transactions on
  Intelligent Transportation Systems}, vol.~19, no.~2, pp. 416--425, 2018.

\bibitem{yuan_yao_e5902fb3}
Y.~Yao, X.~Zhou, and K.~Zhang, ``Density-aware rate adaptation for vehicle
  safety communications in the highway environment,'' \emph{IEEE Communications
  Letters}, vol.~18, no.~7, pp. 1167--1170, 2014.

\bibitem{chuan_xu_39cb2332}
C.~Xu, Z.~Xiong, Z.~Han, G.~Zhao, and S.~Yu, ``Link reliability-based adaptive
  routing for multilevel vehicular networks,'' \emph{IEEE Transactions on
  Vehicular Technology}, vol.~69, no.~10, pp. 11\,771--11\,785, 2020.

\bibitem{alejandro_cohen_cafa02ff}
A.~Cohen, G.~Thiran, V.~B. Bracha, and M.~Médard, ``Adaptive causal network
  coding with feedback for multipath multi-hop communications,'' \emph{IEEE
  Transactions on Communications}, vol.~69, no.~2, pp. 766--785, 2021.

\bibitem{jiaming_cheng_517c1a49}
J.~Cheng, D.~T. Nguyen, and V.~K. Bhargava, ``Resilient edge service placement
  under demand and node failure uncertainties,'' \emph{IEEE Transactions on
  Network and Service Management}, vol.~21, no.~1, p. 558–573, Jun. 2023.

\bibitem{forough_yaghoubi_e705edef}
F.~Yaghoubi, M.~Furdek, A.~Rostami, P.~Öhlén, and L.~Wosinska, ``Design and
  reliability performance of wireless backhaul networks under weather-induced
  correlated failures,'' \emph{IEEE Transactions on Reliability}, vol.~71,
  no.~2, pp. 616--629, 2022.

\bibitem{zhiyu_liu_8d6957dc}
Z.~Liu, B.~Wu, J.~Dai, and H.~Lin, ``Distributed communication-aware motion
  planning for networked mobile robots under formal specifications,''
  \emph{IEEE Transactions on Control of Network Systems}, vol.~7, no.~4, pp.
  1801--1811, 2020.

\bibitem{mohammed_laroui_a3047bc1}
M.~Laroui, A.~Sellami, B.~Nour, H.~Moungla, H.~Afifi, and S.~B. Hacene,
  ``Driving path stability in vanets,'' in \emph{2018 IEEE Global
  Communications Conference (GLOBECOM)}.\hskip 1em plus 0.5em minus 0.4em\relax
  IEEE Press, 2018, p. 1–6.

\bibitem{nikoletta_sofra_c3fa0469}
N.~Sofra, A.~Gkelias, and K.~K. Leung, ``Route construction for long lifetime
  in vanets,'' \emph{IEEE Transactions on Vehicular Technology}, vol.~60,
  no.~7, pp. 3450--3461, 2011.

\bibitem{anirudh_paranjothi_116e5301}
A.~Paranjothi, M.~S. Khan, R.~Patan, R.~M. Parizi, and M.~Atiquzzaman,
  ``Vanetomo: A congestion identification and control scheme in connected
  vehicles using network tomography,'' \emph{Computer Communications}, vol.
  151, pp. 275--289, 2020.

\bibitem{min_li_7968a175}
M.~Li, Z.~Gu, Y.~Long, X.~Shu, Q.~Rong, Z.~Ma, and X.~Shao, ``Retracted: W-gpcr
  routing method for vehicular ad hoc networks,'' \emph{Sensors}, vol.~20,
  no.~12, 2020.

\bibitem{yutong_liu_4f29ab21}
Y.~Liu, K.~Shi, G.~Xu, S.~Lin, and S.~Li, ``Analysis of packet loss
  characteristics in vanets,'' in \emph{2018 8th International Conference on
  Electronics Information and Emergency Communication (ICEIEC)}, 2018, pp.
  219--222.

\bibitem{azzedine_boukerche_3b82d91d}
A.~Boukerche, C.~Rezende, and R.~W. Pazzi, ``A link-reliability-based approach
  to providing qos support for vanets,'' in \emph{2009 IEEE International
  Conference on Communications}, 2009, pp. 1--5.

\bibitem{muddassar_hussain_4948f43c}
M.~Hussain, M.~Scalabrin, M.~Rossi, and N.~Michelusi, ``Mobility and
  blockage-aware communications in millimeter-wave vehicular networks,''
  \emph{IEEE Transactions on Vehicular Technology}, vol.~69, no.~11, pp.
  13\,072--13\,086, 2020.

\bibitem{noura_aljeri_d917ae11}
N.~Aljeri and A.~Boukerche, ``Mobility management in 5g-enabled vehicular
  networks: Models, protocols, and classification,'' \emph{ACM Comput. Surv.},
  vol.~53, no.~5, Sep. 2020.

\bibitem{ilora_maity_ae715f72}
I.~Maity, R.~Dhiman, and S.~Misra, ``Mobiplace: Mobility-aware controller
  placement in software-defined vehicular networks,'' \emph{IEEE Transactions
  on Vehicular Technology}, vol.~70, no.~1, pp. 957--966, 2021.

\bibitem{mao_ye_30a7a94b}
M.~Ye, L.~Guan, and M.~Quddus, ``Tdmp: Reliable target driven and mobility
  prediction based routing protocol in complex vehicular ad-hoc network,''
  \emph{Vehicular Communications}, vol.~31, no.~C, Oct. 2021.

\bibitem{nikki_levering_c8a7d8b9}
N.~Levering, M.~Boon, and M.~Mandjes, ``Estimating probability distributions of
  travel times by fitting a markovian velocity model,'' vol.~24, no.~11, p.
  12372–12392, Jul. 2023.

\bibitem{tarik_taleb_4e090d42}
T.~Taleb, E.~Sakhaee, A.~Jamalipour, K.~Hashimoto, N.~Kato, and Y.~Nemoto, ``A
  stable routing protocol to support its services in vanet networks,''
  \emph{IEEE Transactions on Vehicular Technology}, vol.~56, no.~6, pp.
  3337--3347, 2007.

\bibitem{hamdy_h__el_sayed_48e78f4e}
H.~H. El-Sayed, A.~Younes, and F.~A. Alghamdi, ``Multiobjective multicast dsr
  algorithm for routing in mobile networks with cost, delay, and hop count,''
  \emph{Complexity}, vol. 2021, no.~1, p. 9965872, 2021.

\bibitem{anqi_pan_532ae49f}
A.~Pan, C.~Wang, B.~Shen, and L.~Wang, ``A robust performance evaluation
  approach for solution preservation in multiobjective optimization,''
  \emph{Complex \& Intelligent Systems}, vol.~9, no.~2, pp. 1913--1927, 2023.

\bibitem{ke_li_f6a98174}
K.~Li, R.~Chen, and X.~Yao, ``A data-driven evolutionary transfer optimization
  for expensive problems in dynamic environments,'' \emph{IEEE Transactions on
  Evolutionary Computation}, vol.~28, no.~5, pp. 1396--1411, 2024.

\bibitem{highDdataset}
R.~Krajewski, J.~Bock, L.~Kloeker, and L.~Eckstein, ``The highd dataset: A
  drone dataset of naturalistic vehicle trajectories on german highways for
  validation of highly automated driving systems,'' in \emph{2018 21st
  International Conference on Intelligent Transportation Systems (ITSC)}, 2018,
  pp. 2118--2125.

\end{thebibliography}

	\vspace{0pt}
	\begin{IEEEbiography}[{\includegraphics[width=1in,height=1.25in,clip,keepaspectratio]{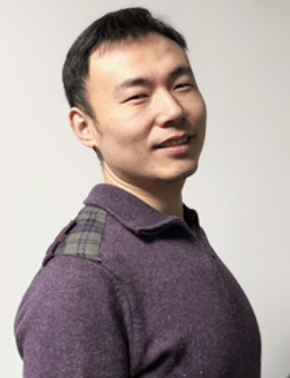}}]{Weian GUO} received the M.Eng. degree in navigation, guidance, and control from Northeastern University, Shenyang, China, in 2009, and the doctor of engineering degree from Tongji University, Shanghai, China, in 2014. From 2011 to 2013, he was sponsored by China Scholarship Council to carry on his research at the Social Robotics Laboratory, National University of Singapore. He is currently an associate Professor with the Sino-German College of Applied Science, Tongji University. His interests include computational intelligence, control theory and vehicular network.
	\end{IEEEbiography}
	\vspace{-50pt}
	\begin{IEEEbiography}[{\includegraphics[width=1in,height=1.25in,clip,keepaspectratio]{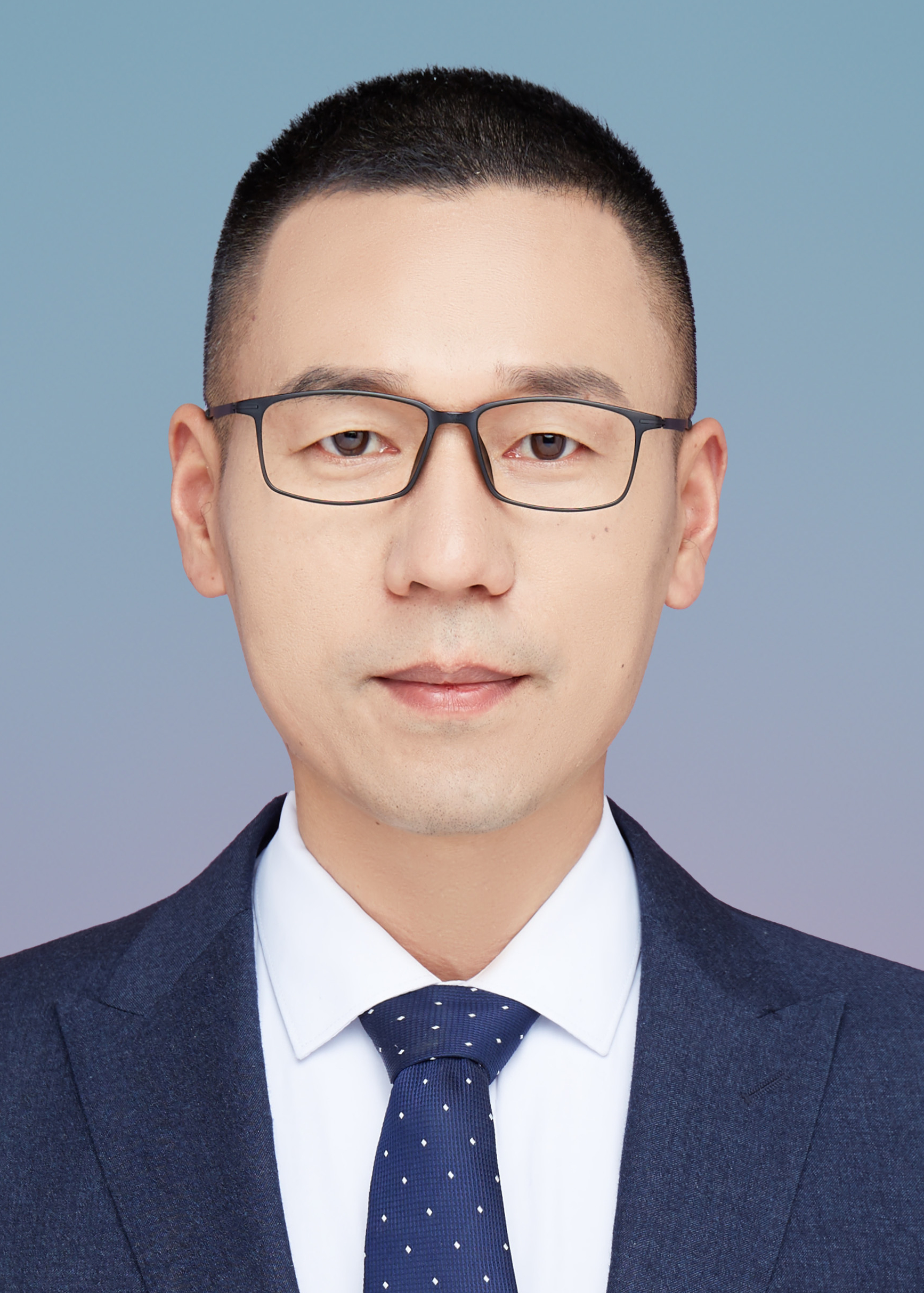}}]{Wuzhao LI} received the Ph.D. degree in Control Science and Engineering from Tongji University, Shanghai, China. In 2022, he was a Professor with the Wenzhou Polytechnic, China. He has authored or coauthored more than 20 papers in international SCI journals. His research interests include machine learning, Multi-Objective optimization and robust control.
	\end{IEEEbiography}
	\vspace{-50pt}
	\begin{IEEEbiography}[{\includegraphics[width=1in,height=1.25in,clip,keepaspectratio]{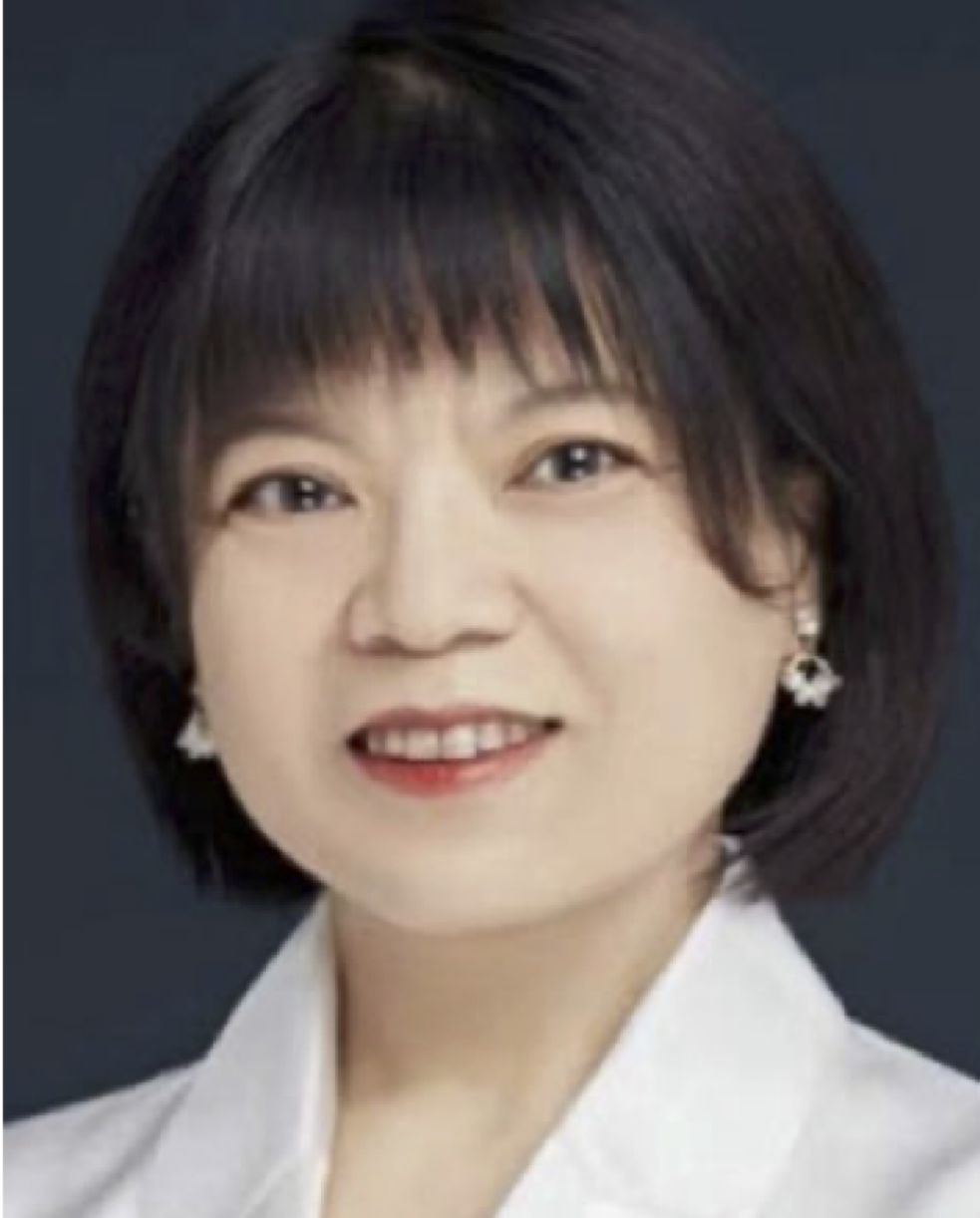}}]{Li LI} received the B.S. and M.S. degrees in electrical automation from Shengyang Agriculture University, Shengyang, China, in 1996 and 1999, respectively, and the Ph.D. degree in mechatronics engineering from the Shenyang Institute of Automation, Chinese Academy of Science, Shenyang, in 2003. She joined Tongji University, Shanghai, China, in 2003, where she is currently a professor of control science and engineering. She has over 50 publications, including 4 books, over 30 journal papers, and 2 book chapters. Her current research interests include production planning and scheduling, computational intelligence, data-driven modeling and optimization, semiconductor manufacturing, and energy systems.
	\end{IEEEbiography}
	\vspace{-50pt}
	\begin{IEEEbiography}[{\includegraphics[width=1in,height=1.25in,clip,keepaspectratio]{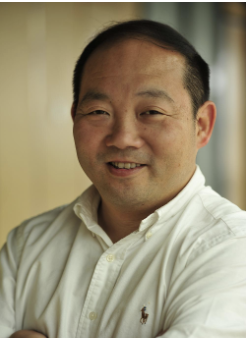}}]{Lun ZHANG}
		received the B.S. and Ph.D degrees in computer communications, transportation information engineering and control, Tongji University, Shanghai, China, in 1992 and 2005, respectively. He is currently a professor with School of Transportation, Tongji University. His interests include intelligent transportation, computational intelligence, and deep learning.
	\end{IEEEbiography}	
	\vspace{-50pt}
	\begin{IEEEbiography}[{\includegraphics[width=1in,height=1.25in,clip,keepaspectratio]{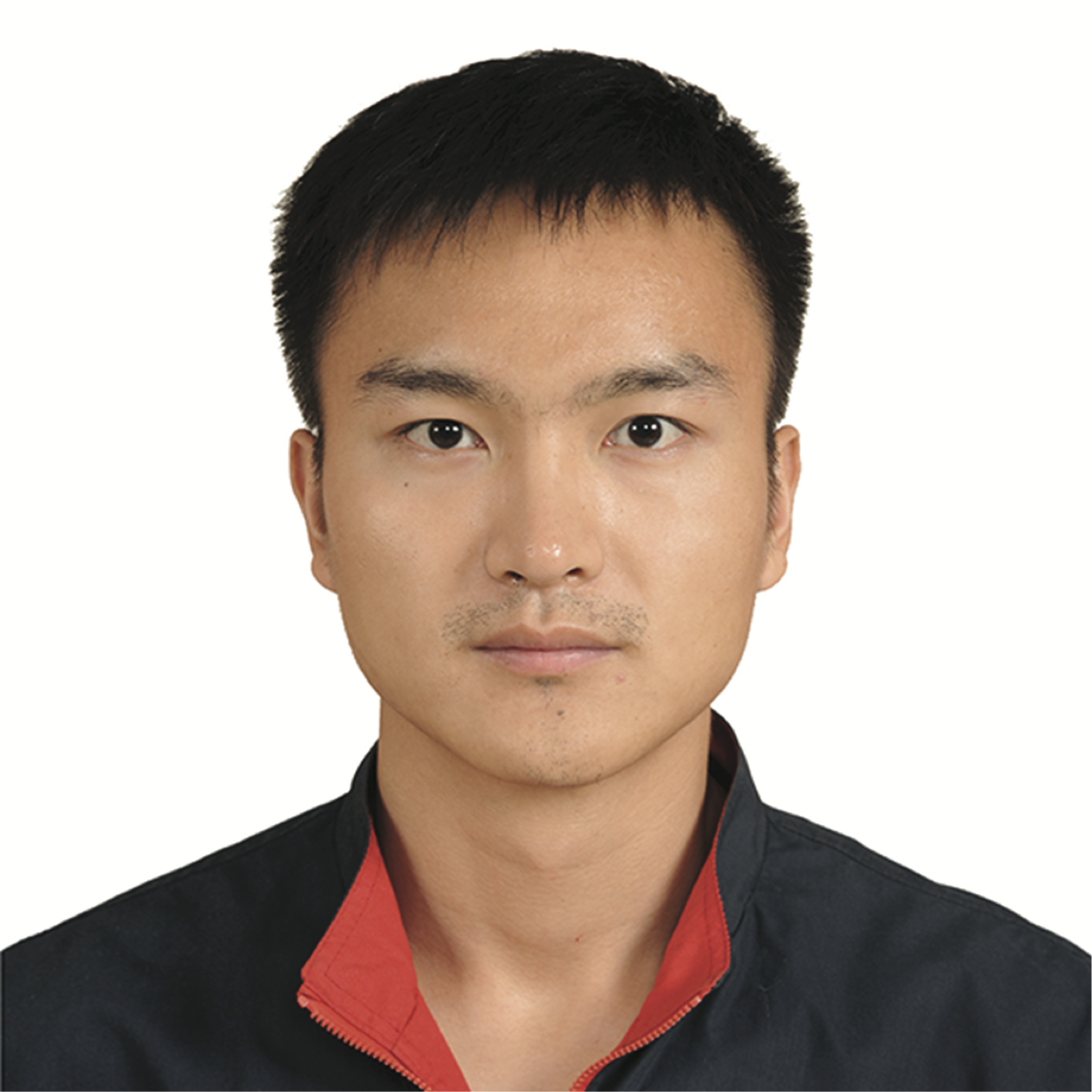}}]{Dongyang Li}
		received the M.S. and Ph.D degrees in school of electronics and information engineering, Tongji University, Shanghai, China, in 2017 and 2022, respectively. From 2019 to 2021, he was sponsored by China Scholarship Council to carry on his research at Georgia Institute of Technology. He is now an engineer with the Sino-German College of Applied Science, Tongji University; His research interest includes computational intelligence, deep learning and their applications.
	\end{IEEEbiography}
	
\end{document}